\pdfoutput=1

\documentclass[11pt]{article}

\usepackage{acl}

\usepackage{times}
\usepackage{latexsym}

\usepackage[T1]{fontenc}

\usepackage[utf8]{inputenc}

\usepackage{microtype}

\usepackage{inconsolata}

\usepackage{graphicx}
\usepackage{xcolor}
\usepackage{booktabs}
\usepackage{amsthm, amsmath, amssymb}
\usepackage{mathtools}
\usepackage{graphicx,subcaption,float}
\usepackage{breqn}
\usepackage[autostyle]{csquotes}
\usepackage[british]{babel}
\usepackage{breakcites}
\usepackage{bbm}
\usepackage{multirow}
\usepackage{hyperref}
\usepackage{multirow}
\usepackage{comment}

\usepackage{array}
\newcolumntype{x}[1]{>{\centering\arraybackslash\hspace{0pt}}p{#1}}

\usepackage{float}
\usepackage{colortbl}
\usepackage{xcolor}

\usepackage{bbold}

\usepackage{pgfplots}
\pgfplotsset{compat=1.8}
\makeatletter
\pgfarrowsdeclare{DIN}{DIN}
{
  \pgfutil@tempdima=0.5pt%
  \advance\pgfutil@tempdima by.25\pgflinewidth%
  \pgfutil@tempdimb=7.29\pgfutil@tempdima\advance\pgfutil@tempdimb by.5\pgflinewidth%
  \pgfarrowsleftextend{+-\pgfutil@tempdimb}
  \pgfutil@tempdimb=.5\pgfutil@tempdima\advance\pgfutil@tempdimb by1.6\pgflinewidth%
  \pgfarrowsrightextend{+\pgfutil@tempdimb}
}
{
  \pgfutil@tempdima=0.5pt%
  \advance\pgfutil@tempdima by.25\pgflinewidth%
  \pgfsetdash{}{+0pt}
  \pgfsetmiterjoin
  \pgfpathmoveto{\pgfpointadd{\pgfqpoint{0.5\pgfutil@tempdima}{0pt}}{\pgfqpoint{-4mm}{0.5mm}}}
  \pgfpathlineto{\pgfqpoint{0.5\pgfutil@tempdima}{0\pgfutil@tempdima}}
  \pgfpathlineto{\pgfpointadd{\pgfqpoint{0.5\pgfutil@tempdima}{0pt}}{\pgfqpoint{-4mm}{-0.5mm}}}
  \pgfpathclose
  \pgfusepathqfillstroke
}
\pgfarrowsdeclarereversed{DIN reversed}{DIN reversed}{DIN}{DIN}

\def\bx{\mathbf{x}}

\def\bp{\mathbf{p}}

\def\bX{\mathbf{X}}

\def\bL{\mathbf{L}}

\def\b0{\mathbf{0}}
\def\b1{\mathbf{1}}

\def\cL{\mathcal{L}}

\DeclareMathOperator*{\argmax}{arg\,max}

\newcommand{\sqrbrkt}[1]{\{#1\}}

                    \newcommand{\cv}{\mathcal{V}}

\newcommand{\newcrossmark}{\scalebox{0.85}[1]{$\times$}}

\title{Exploring Ordinality in Text Classification: \\ A Comparative Study of Explicit and Implicit Techniques}

\author{Siva Rajesh Kasa* \\
  Amazon, India \\
  \texttt{kasasiva@amazon.com} \\\And
  Aniket Goel* \\
  IIIT Dehli\\
  \texttt{aniket20281@iiitd.ac.in} \\\And
  Karan Gupta* \\
  Amazon, India\\
  \texttt{karaniis@amazon.com} \\\AND
  Sumegh Roychowdhury* \\
  Amazon, India\\
  \texttt{sumegr@amazon.com} \\\And
  Anish Bhanushali \\
  Amazon, India\\
  \texttt{ankeeb@amazon.com} \\\And
  Nikhil Pattisapu \\
  Amazon, India\\
  \texttt{npattisa@amazon.com} \\\And
  Prasanna Srinivasa Murthy \\
  Amazon, India\\
  \texttt{sprsn@amazon.com} \\
}
\begin{document}

\maketitle
\setlength{\abovedisplayskip}{3pt}
\setlength{\belowdisplayskip}{3pt}
\def\thefootnote{*}\footnotetext{These authors contributed equally to this work}\def\thefootnote{\arabic{footnote}}
\begin{abstract}

Ordinal Classification (OC) is a widely encountered challenge in Natural Language Processing (NLP), with applications in various domains such as sentiment analysis, rating prediction, and more. Previous approaches to tackle OC have primarily focused on modifying existing or creating novel loss functions that \textbf{explicitly} account for the ordinal nature of labels. 
However, with the advent of Pretrained Language Models (PLMs), it became possible to tackle ordinality through the \textbf{implicit} semantics of the labels as well. 
This paper provides a comprehensive theoretical and empirical examination of both these approaches.
Furthermore, we also offer strategic recommendations regarding the most effective approach to adopt based on specific settings.

\end{abstract}

\section{Introduction}

Ordinal classification (OC) is a key task in natural language processing (NLP), with many applications that require ordering or ranking in the output
such as Sentiment Analysis \cite{dang2020sentiment}, Rating Prediction \cite{liu2020yelp}, Age Group Classification \cite{sanchez2022age}, etc. In each of these examples, the output categories have a natural ordering, making these tasks suited to ordinal classification.

Broadly speaking, OC task can be tackled using one of the following two approaches -  \textbf{explicit} vs \textbf{implicit}. The classical explicit approach relies on tweaking the loss function based on an
\textit{explicit notion} of distance between labels and penalising based on the degree of misclassification \citep{castagnos2022simple,diaz2019soft}.
Alternatively, the second \textit{implicit} approach we propose is based on more recent advancements in language modeling. This approach organically engages the semantics of the labels, thereby harnessing their inherent characteristics for the classification task. Thus, while the former represents a time-tested approach, the latter offers a new way to tackle OC, bringing the richness of semantic information into the classification process. Our paper comprehensively studies the different techniques that fall under both these strategies.
The classical approach relies on tweaking Cross Entropy (CE) which is a commonly used loss function for nominal  classification (NC) (where the assumption is that the classes are mutually exclusive and have no inherent order or relationship to each other) \citep{yamasaki2022unimodal,castagnos2022simple,diaz2019soft}. The performance in NC task is usually measured in terms of accuracy-based metrics such F1 scores. While CE is optimal for NC, since it treats all misclassifications as the same, it is sub-optimal for OC. There are several approaches/tweaks proposed to extend CE loss for OC, such as the ones proposed by \citet{diaz2019soft,castagnos2022simple}, which adds a penalty based on the absolute difference in the class rankings. These loss
functions are designed such that the more the distance between prediction and ground truth, the more the penalty. The performance of these OC tasks is measured in terms of Mean Squared Error (MSE), Mean Absolute Error (MAE) and Off-by-1 (OB1) accuracy \cite{castagnos2022simple}. 

Our work fills a crucial gap in the current understanding of these losses used in OC. Prior to our research, a unifying analysis of these different loss based approaches was conspicuously absent, making it challenging to holistically compare and contrast their characteristics and performance. In an attempt to address this lacuna, we evaluate these loss functions for four desirable properties, namely, proper scoring rule, unimodality, convexity and ordinality. We also propose a loss function which exhibits these desirable properties while demonstrating that several of these loss functions can be interpreted as specialized versions of the proposed generalized loss function.

Pretrained Language Models (PLMs), which include both encoder models (like BERT \cite{devlin-etal-2019-bert}) and decoder models (like GPT \cite{radford2018improving}), offer a robust mechanism to learn the semantics or representations of words. These models are trained on vast amounts of text data, during which they develop an understanding of the context in which words are used. Encoder models like BERT learn to predict a word based on its surrounding context, thereby creating a rich and nuanced understanding of word semantics. On the other hand, decoder models like GPT generate a sequence of words, learning to predict the next word in a sentence, hence understanding semantics in a left-to-right context manner. This learning process enables these models to develop high-dimensional vector representations (embeddings) that capture the underlying semantics of words.

Interestingly, these semantically rich embeddings can be utilized to \textit{implicitly} factor in the ordinality of labels in OC tasks. The numeric vector representation of words inherently carries semantic relationships that can mirror ordinal relationships. For instance, the embeddings of words like "good", "neutral", and "bad" reflect their comparative semantics in the vector space. When used for OC tasks, such as sentiment analysis or rating prediction, these learned embeddings can provide a more natural and effective way to understand and encode the ordinality of labels. The model does not just see these labels as distinct classes, but as points on a scale, thus allowing a more nuanced approach to such classification tasks.

In our study, we make the following contributions to advance the understanding and application of OC in NLP:
\begin{itemize}
    \item We conduct a comprehensive analysis of several explicit loss-based approaches for Ordinal Classification (OC), examining them through the lens of useful theoretical properties, such as proper scoring rule, convexity, unimodality and ordinality. Our empirical findings demonstrate that previously proposed techniques for solving OC excel primarily in ordinal metrics while compromising performance on nominal metrics. This analysis has led us to propose a hybrid loss function that achieves a better balance between nominal and ordinal metrics compared to the performance of its individual components.

    \item Further, we study two general PLM based methodologies (encoder \& decoder-based) that implicitly factor in ordinality. These approaches signify a paradigm shift from traditional methods, opening up new avenues in ordinal classification.

    \item Lastly, we undertake an exhaustive comparison of these explicit and implicit approaches under different scenarios, providing necessary ablations and conclude by offering strategic recommendations on the suitable choice between these approaches in \S\ref{sec:conclusion}.
\end{itemize}

The rest of the paper is organized as follows: In \S \ref{sec:explicit_intro}, we discuss the various explicit approaches i.e. review the loss functions and discuss their theoretical properties. Using this as motivation, we also propose a new hybrid loss function and study it's properties.
In \S \ref{sec:implicit_entailment_intro}, we discuss the encoder models and show how OC can be approached from an entailment-style modeling perspective. In \S \ref{sec:implicit_decoder_intro}, we discuss the decoder models and how OC can be approached from a next word prediction task perspective. Figure~\ref{fig:main_figure} depicts the explicit as well as implicit approaches. In \S\ref{sec:results_discussion}, we empirically compare these three methods and give our recommendations.

\input{main_figure}

\section{Explicit approach: Loss-functions and Analysis} \label{sec:explicit_intro}

\begin{table}[ht]
\scriptsize
    \centering
    \begin{tabular}{c|c|c|c|c}
        Loss & PSR & UM & CX & Ord \\ \hline
         CE & $\checkmark$ & \newcrossmark &  $\checkmark$ & low \\
         OLL & $\checkmark$ & \newcrossmark &  $\checkmark$ & high \\
         MLL & $\checkmark$ & \newcrossmark &  $\checkmark$ & high \\
         SOFT & \newcrossmark & \newcrossmark &  $\checkmark$ & low \\
         EMD & $\checkmark$ & \newcrossmark &  $\checkmark$ & low \\
         CORAL & \checkmark & \newcrossmark &  $\checkmark$ & low \\
         VS-SL & \newcrossmark & $\checkmark$ &  \newcrossmark & low \\
         WKL & $\checkmark$ & \newcrossmark &  \newcrossmark & low \\
    \end{tabular}
    \caption{Various properties satisfied by different loss based approaches. \textit{Notation}: PSR - Proper Scoring Rule, UM - Unimodality, CX - Convexity, Ord - Ordinality}
    \label{tab:summary_loss_properties}
\end{table}

Let $\{(\bX_i,y_i)\}_{i=1}^{N}$ be $N$ independent and identically distributed datapoints containing the input features $\bX_i$ and their corresponding labels $y_i$; where $y_i \in \{1,\dots,K\}$ where $K$ is the number of classes. The output of the classifier is denoted by $\Phi_{\theta}(\bX_i) = \hat{\bp}_i = (\hat{p_{i_1}},\dots,\hat{p_{i_K}})$ which is a probability distribution over the $K$ classes.
Let $\mathbb{I}(y_i)$ be the one hot encoding of $y_i$.
The classifier is trained by optimizing the parameters $\theta$  such that  $\frac{1}{N}\sum_{i=1}^{N}\cL_{\theta}(\hat{\bp_i},\mathbb{I}(y_i))$ reaches a minimum. In the rest of the paper, for ease of mathematical exposition, we omit the indexing with respect to $i$, i.e. remove the $\frac{1}{N}$ and $\sum_{i=1}^{N}$, wherever it is evident from the loss expression.

In the next subsection, we give a few desirable theoretical properties of the loss function $\cL$ in the context of OC and then follow it up by discussing some of the widely used loss functions.

\subsection{Desirable Properties of Losses in OC}

\textbf{\textit{Proper Scoring Rule (PSR):}} A loss is said to be a PSR \citep{gneiting2007strictly,merkle2013choosing} if it takes the lowest value when the predicted class probabilities match the ground truth which is a one-hot encoded $K$-dimensional vector.
Being PSR ensures that the loss indeed tries to optimize the classifier to predict the ground truth without injecting any bias in the predicted outputs. Further, PSR losses also help to produce well calibrated probabilities \citep{lakshminarayanan2017simple}.

\textbf{\textit{Convexity (Cx):}} Convexity of $\cL$ with respect to $\hat{\bp}$ is a desirable property of the loss as it is an essential requirement for several convex formulations of Neural Networks (NNs) \cite{kawaguchi2019every,du2019gradient,pilanci2020neural,wojtowytsch2023stochastic}. Further, if both $\cL$ and $\Phi$ (classifier function) are convex with respect to $\theta$ (as in the case of logistic regression or support vector machines), then it is guaranteed that the local minima indeed coincides with the global minima.

Several widely used losses such CE, MAE, MSE, etc. are both PSRs and Convex. Next, we look at two desirable characteristics of the output probabilities from the classifier in the context of OC. 

\textbf{\textit{Unimodality (UM):}} If the output probabilities have single mode, i.e. $\hat{p_j} > \hat{p_l} < \hat{p_i}$ is not satisfied for any $1 \leq j < l < i \leq K$, then we say the classifier satisfies the UM condition \citep{beckham2017unimodal,yamasaki2022unimodal,iannario2011cub}. An illustration is given in Figure \ref{fig:unimodality}.

\textbf{\textit{Ordinality (Ord):}} In the context of OC, we require the loss $\cL$ to explicitly penalize the mis-classifications more which are farther away from each other compared to the ones which are closer. The goal is to enforce a meaningful ordering among the labels unlike nominal classifcation where categories lack a specific order. We will see later that different loss functions enforce ordinality to a varying degree. 

\noindent In the next subsection, we discuss some of the widely used loss functions in the context of OC in NLP. 
\subsection{Widely used Loss Functions in OC}

\textbf{\textit{Cross-Entropy (CE):}} CE is given by 
\begin{align}
- \sum_{k=1}^K {p_k} \log(\hat{p}_{k})  \label{eq:crossentropy} 
\end{align}
The above expression boils down to $-\log(\hat{p}_{y_i}) $. Oftentimes CE is discredited for not being able to factor in ordering as its expression does not take into account the probabilities corresponding to the non-groundtruth classes.
\textbf{\textit{Ordinal Log Loss (OLL) \citep{castagnos2022simple}:}} OLL is given by

\begin{align}
    - \sum_{k \neq y} |k - y|^{\alpha} \log(1- \hat{p}_{k}). \label{eq:OLL}
\end{align}

Here $\alpha > 0$ is a hyperparameter. This can be seen as a complementary of CE where the missclassification is explicitly penalized in proportion to its degree by the term 
$|k -y|^{\alpha}$. Both CE and OLL belong to PSR family as both these losses are zero when the output probabilities coincide with the ground truth one-hot encoding.

\textbf{\textit{SOFT labels (SOFT) \cite{diaz2019soft}:}} In this approach, first the ground truth $K$-dimensional onehot encoded is modified to $K$ `soft' labels as follows. 
\begin{align}
    p_{k}^{\text {soft }}=\frac{\exp (-\beta |k - y|)}{\sum_{k'} \exp (-\beta |k'- y|)} \label{eq:softlabels}
\end{align}

Then CE loss can be computed using these soft ground labels as truth probabilities. 
\begin{align}
    - \sum_{k=1}^K p_{k}^{\text {soft }}\log(\hat{p}_{k}) \label{eq:softlabels_loss}
\end{align}

Clearly, introducing softlabels makes this approach not fall in the PSR family as the ground-truth onehot encoding vector does not minimize the loss anymore.

\textbf{\textit{Earth Mover Distance (EMD) \cite{rubner2000earth} :}} EMD  or Wasserstein loss is defined 
$$\text{EMD} = (\text{CDF}(\mathbb{I}(y_i)) - \text{CDF}(\hat{\bp_i}))^2$$

Here CDF refers to the cumulative distribution function. While EMD is a PSR, it does not impose a strong penalty in the tails (as compared to OLL) because CDFs are monotonic with range in $[0,1]$ and hence the difference between the CDFs will be small in the tails.

\textbf{\textit{Unimodal Losses:}} \citealt{da2008unimodal} and \citealt{beckham2017unimodal} introduce a parametric way to force unimodality in the predicted probability scores by first computing a scalar function $f(\bx) \in [0,1]$ and using this scalar as the parameter of a Binomial distribution with parameters $ (K,f(\bx))$, the Probability Mass Function (PMF) is computed which forms the final predicted probabilities.
Note that the computing a single scalar $f(x)$  (as opposed a vector of embeddings as done in other methods) severely hampers the learning; thus the guarantee of unimodality comes at the cost of degradation in performance. To address this, \citet{yamasaki2022unimodal} proposes a more-flexible unimodal OC framework by imposing shape-based constraints on the output probabilities; the V-Shaped Stereotyped Logit (VS-SL) method has shown to be the state-of-the-art for UM in their work and hence, we use it as a baseline in our experiments.  

We also include two more loss function variants, namely COnsistent RAnk Logits (CORAL) \cite{Cao_2020} and Weighted Kappa Loss (WKL) \cite{article}, and benchmark their effectiveness in terms of accuracy and ordinality.

To benchmark the performance of the above loss functions, we run experiments on 
three
benchmark multi-class text classification datasets,  each with its distinct task: \textit{Hypothesis entails Premise} task - SNLI \cite{bowman-etal-2015-large}, \textit{Reviews Classification} task - Amazon Reviews (AR) \cite{keung-etal-2020-multilingual}, 
and \textit{Sentiment Analysis} task - SST-5 \cite{socher-etal-2013-recursive}. We report weighted-F1 scores and MAE, MSE, Off-by-1 (OB1) accuracy to measure both classification and ordinal performance respectively. In the interest of space, we present the details of datasets and metrics in Appendix \ref{app:datasets} and \ref{app:metrics}.

We observe that in general CE performs best in terms of nominal metrics (like weighted-F1) and OLL performs best in terms of ordinal metrics on an average. However, there seems to be a trade-off between nominal and ordinal performance i.e. the improvement in ordinal metrics comes at the expense of nominal metrics. However, given that all these metrics are also from the PSR family, and in some sense are not independent of each other as a perfect classifier would improve both these metrics simultaneously. 
This intuition prompted us to experiment with a weighted combination of CE and OLL, which we have named Multitask Log Loss (MLL), hypothesizing that it would inherit the best attributes of both methods.

\textbf{\textit{Multi-task log loss function (MLL):}} MLL is given by 
\begin{align}
MLL  = \lambda \times CE + (1 - \lambda) \times OLL 
\end{align}
Here, $\lambda \in [0,1]$ is a hyperparameter. MLL satisfies both convexity and PSR conditions. Further, while OLL and MLL are not theoretically guaranteed to be UM, but empirically have been found to be satisfying UM condition for 80-90\% test datapoints which we show later in Figure~\ref{fig:unimodal}. A summary of the properties satisfied by different loss-based approaches is given in Table \ref{tab:summary_loss_properties}, with discussions in Appendix \ref{app:psr_proofs} and \ref{app:convexity_proofs}. Note that while it's theoretically possible to consider weighted combinations involving other loss functions, we only considered CE and OLL here because OLL has already been shown to outperform other losses in terms of ordinal metrics \citep{castagnos2022simple} and we aimed to improve its performance in term of nominal metrics as well, by adding with the CE term.

We perform all explicit approach based experiments using BERT-base (refer Appendix~\ref{sec:train_detail}). Following \citet{castagnos2022simple} we also train a smaller version - TinyBERT \cite{jiao-etal-2020-tinybert}, as the performance of different loss functions
is better contrasted when the size of the base model is small, which is usually the case in online settings where we cannot deploy larger models.
The detailed results are given in Table~\ref{tab:tinybert_main} and Appendix~\ref{app:loss_comparison}.

\section{Implicit approach: Entailment-style Encoder Models} \label{sec:implicit_entailment_intro}
\citet{wang2021entailment} proposed reformulating vanilla classification into an entailment-style task to enhance the few-shot capabilities of PLMs. Here, the model learns to predict whether the input text and the label entail each other or not (similar to Natural Language Inference (NLI) setting), leveraging the inherent semantic relationship between the label and input text. We adopt a similar approach for our task and explore it through the lens of ordinality, which has not been studied in prior works.

We assume the existence of a classifier, based on Pretrained Language Model (PLM), called $\Phi_{\theta}$.
Let $\bL = \{L_1, L_2, \dots, L_K\}$ be the collection of textual labels.
The training dataset can be divided based on the labels into $\mathcal{D}_{\text{tr}}$, which consists of subsets ${D_1, D_2, \dots, D_K}$. Each subset contains the available training data $\left\{\bx_{k_i}\right\}_{i=1}^{n_k}$ for label $L_k$. The corresponding test data is represented by $\mathcal{D}_{\text{tst}}$.

During the training phase in $\mathcal{D}_{\text{tr}}$, the following entailment-style data augmentation technique is employed:
For each data point $\bx_{i_j}$ with a ground truth label of $L_j$, $K$ samples ($\{s_{i_j}^k\}_{k=1}^{K}$) are generated and augmented as follows:
\begin{equation}
     s_{i_j}^{k} = \{\bx_{i_j} + verbaliser(L_{k}),\mathbb{1}_{j}(k) \}_{k=1}^{K}
\label{eq:entail}
\end{equation}

Here, $\mathbb{1}_j(k)$ is an indicator function that yields 1 if $k=j$ else 0. The `+' operator denotes concatenation operation (refer \S\ref{sec:train_detail}), and $verbaliser()$ is a pre-defined template (specific to the downstream task) describing the label in natural language. For example, in sentiment classification task $verbaliser(L_{j} = positive)$ can be described as: \texttt{indicates \textit{positive} sentiment}. See Figure~\ref{fig:main_figure} for example.

Essentially, for each data point, $(K-1)$ negative samples and 1 positive sample are created. Finally the problem reduces to the following NLI task - Does $\bx_{i_j}$ entail $verbaliser(L_j)$ or not?
Once these $K \times \sum n_k$ augmented examples are generated, the parameters $\theta$ are finetuned for a binary classification task, where $s_{i_j}^k$ serves as the input and $\mathbb{1}_j(k)$ acts as the ground truth.
During the inference phase, for a datapoint $\bx$ the predicted label $\hat{L}$ is obtained using:
\begin{equation}
    \hat{L} = \argmax_k \{ \Phi_{\theta}(s^{k}) \} 
    \ \forall \ i \in \sqrbrkt{1,\dots,n_k}
\end{equation}
During inference for $\bx$, $s^{k}$ is computed following Eq.~\ref{eq:entail} and \texttt{softmax()} is applied on the predicted logits before taking \texttt{argmax} so that all the class probabilities sum up to 1. As the model leverages the natural language meanings of the labels during training, we argue it is inherently capable of learning to predict labels that are \textbf{ordinally consistent}. For instance, the model learns to comprehend that the label \texttt{very negative sentiment} is closer in semantic space to \texttt{negative sentiment} than to \texttt{very positive sentiment}. This understanding prevents the model from deviating significantly from the actual ground truth. In contrast, in the case of vanilla CE, these labels are treated solely as numbers, disregarding their inter-semantic relations. We again use BERT-base here as base model for performing experiments. The exact label verbalisers used for all datasets are mentioned in Appendix~\ref{table:prompt_details}.

\section{Implicit approach: Generative Models}
\label{sec:implicit_decoder_intro}

\begin{table*}[ht]
\begin{minipage}[t]{.22\linewidth}
\tiny
\centering
\addtolength{\tabcolsep}{-3.8pt}  
\begin{tabular}[t]{llx{0.48cm}x{0.48cm}}
&     & \multicolumn{2}{c}{100 \%} \\
Dataset        & Metric & CE           & MLL                    \\ \hline
               \multirow{4}{*}{\rotatebox{90}{ \parbox{1.5cm}{SNLI} }}           & F1     & 0.821 (0.00) & 0.832 (0.00)  \\
               & MSE    & 0.264 (0.00) & 0.257 (0.01)  \\
               & MAE    & 0.208 (0.02) & 0.205 (0.01)  \\
               & OB1    & 0.972 (0.01) & 0.974 (0.00)    \\ \hline  
\multirow{4}{*}{\rotatebox{90}{ \parbox{1.5cm}{SST5} }}           & F1     & 0.357 (0.01) & 0.378 (0.02)  \\
               & MSE    & 1.197 (0.00) & 1.125 (0.01)  \\
               & MAE    & 0.768 (0.02) & 0.742 (0.01)  \\
               & OB1    & 0.852 (0.00) & 0.863 (0.00)  \\ \hline
                \multirow{4}{*}{\rotatebox{90}{ \parbox{1.5cm}{Amazon \newline Reviews} }} & F1     & 0.543 (0.02) & 0.544 (0.00)  \\
               & MSE    & 0.904 (0.04) & 0.819 (0.01)  \\
               & MAE    & 0.581 (0.01) & 0.568 (0.00)  \\ 
               & OB1    & 0.903 (0.00) & 0.915 (0.00)  \\ \hline

\end{tabular}
\caption{\small CE vs MLL using TinyBERT as base model. For full comparison refer Table~\ref{tab:losses-bert-tiny} in Appendix \ref{app:loss_comparison}.}
\label{tab:tinybert_main}
\addtolength{\tabcolsep}{3.8pt}
\end{minipage}
\hspace{0.18cm}%
\begin{minipage}[t]{.73\linewidth}
\tiny
\addtolength{\tabcolsep}{-3.8pt}    
\begin{tabular}[t]{clx{0.48cm}x{0.48cm}x{0.48cm}x{0.48cm}@{\hskip 0.11in}x{0.48cm}x{0.48cm}x{0.48cm}x{0.48cm}@{\hskip 0.11in}x{0.48cm}x{0.48cm}x{0.48cm}x{0.48cm}@{\hskip 0.11in}x{0.48cm}x{0.48cm}x{0.48cm}x{0.48cm}}
 &     & \multicolumn{4}{c}{100 \%} & \multicolumn{4}{c}{50 \%} &   \multicolumn{4}{c}{25 \%} &   \multicolumn{4}{c}{10 \%} \\
  &     & CE   & MLL   & ENT & GPT &   CE   & MLL   & ENT   & GPT   &   CE   & MLL   & ENT  & GPT  &   CE   & MLL   & ENT   & GPT  \\ \hline
 \multirow{4}{*}{\rotatebox{90}{SNLI}} & F1 & 0.890  (0.02) &  \cellcolor{green!=45}0.891 (0.02) & 0.885 (0.30) & 0.776 (0.00)   & 0.873 (0.00) &  0.882 (0.09) & \cellcolor{green!=45}0.885 (0.04) & 0.654 (0.00) &   0.861 (0.00)  & 0.865 (0.04) & \cellcolor{green!=45}0.869 (0.01) & 0.527 (0.01) &  0.836 (0.00) & \cellcolor{green!=45}0.848 (0.04) & 0.845 (0.01) & 0.389 (0.01) \\
 & MAE & 0.123 (0.01) & \cellcolor{green!=45}0.122 (0.02) & 0.128 (0.00) & 0.261 (0.00)  & 0.146 (0.00) & 0.131 (0.01) & \cellcolor{green!=45}0.129 (0.00)  & 0.424 (0.00)  &      0.161 (0.00)    & 0.153 (0.01) & \cellcolor{green!=45}0.150 (0.00)   &  0.599 (0.02)  &   0.191 (0.00) & \cellcolor{green!=45}0.171 (0.05) & 0.181 (0.00)  & 0.811 (0.01)\\
 & MSE & 0.152 (0.04) & \cellcolor{green!=45}0.149 (0.04) & 0.157 (0.01) & 0.338 (0.00)  & 0.188 (0.00) & \cellcolor{green!=45}0.156 (0.05)& 0.159 (0.00) &   0.596 (0.00)  &  0.206 (0.00)     & \cellcolor{green!=30}0.190 (0.00) &  \cellcolor{green!=30}0.190 (0.00)   & 0.878 (0.04)   &    0.249 (0.01)  & 0.210 (0.06)  & \cellcolor{green!=45}0.236 (0.00)  & 1.263 (0.04)\\
 & OB1 & 0.985 (0.01) & \cellcolor{green!=45}0.986 (0.01)  & 0.985 (0.02) & 0.961 (0.00)  & 0.979 (0.00) & \cellcolor{green!=45}0.987 (0.02)      &  0.980 (0.00)  & 0.913 (0.00)    &      0.977 (0.00)     &  \cellcolor{green!=45}0.981 (0.02)    & 0.980 (0.00)   & 0.860 (0.01)      &       0.971 (0.00) & \cellcolor{green!=45}0.980 (0.00) & 0.972 (0.00)  & 0.774 (0.01) \\
\hline
\multirow{4}{*}{\rotatebox{90}{SST5}} & F1  & 0.484 (0.01) & 0.492 (0.01) & \cellcolor{green!=45}0.508 (0.01)  & 0.487 (0.02)    & 0.442 (0.02)    &   0.468 (0.03)    &  0.46 (0.02)      &  \cellcolor{green!=45}0.476 (0.00) &         0.415 (0.03)    & \cellcolor{green!=45}0.440 (0.04) &  0.423 (0.03)     & 0.316 (0.01) &   0.417 (0.02)  & 0.428 (0.08) & \cellcolor{green!=45}0.430 (0.03)  & 0.274 (0.02)\\
 & MAE & 0.576 (0.02) & 0.575 (0.02)  & \cellcolor{green!=45}0.543 (0.01)  & 0.585 (0.01)    & 0.620 (0.01)    &   0.600 (0.02)     &  \cellcolor{green!=45}0.585 (0.01)   & 0.595 (0.00)   &          0.652 (0.04)     &   0.620 (0.04)    & \cellcolor{green!=45}0.619 (0.04)     &  0.966 (0.03)  &      0.697 (0.02) & 0.662 (0.10) & \cellcolor{green!=45}0.651 (0.01)  & 1.043 (0.02) \\
 & MSE & 0.761 (0.04) & 0.757 (0.01) & \cellcolor{green!=45}0.683 (0.02)    & 0.802 (0.02)   & 0.845 (0.04)   & 0.760 (0.04)      &  \cellcolor{green!=45}0.694 (0.04)   &  0.819 (0.01)   &           0.910 (0.10)    &   \cellcolor{green!=45}0.827 (0.12)    & 0.870 (0.01)  & 1.740 (0.10)   &     1.04 (0.06) & 0.905 (0.25) & \cellcolor{green!=45}0.871 (0.01)  & 1.897 (0.19) \\
 & OB1 &   0.925 (0.01)   &    0.931 (0.00)   &  \cellcolor{green!=45}0.932 (0.01)   & 0.918 (0.00)    & 0.910 (0.08) &      0.929 (0.01)      &  \cellcolor{green!=45}0.93 (0.01)   & 0.912 (0.00) &       0.902 (0.01)    &    \cellcolor{green!=45}0.918 (0.02)   & 0.911 (0.02)  & 0.769 (0.01)         &     0.875 (0.01) & 0.898 (0.05) & \cellcolor{green!=45}0.915 (0.02)  & 0.734 (0.01)  \\
\hline
\multirow{4}{*}{\rotatebox{90}{ \parbox{1.5cm}{Amazon \newline Reviews} }} & F1  & 0.586 (0.04) & \cellcolor{green!=45}0.589 (0.04) & 0.585 (0.03)   & 0.522 (0.00)     & 0.573 (0.00)  & 0.579 (0.01) &  \cellcolor{green!=45}0.586 (0.02) & 0.449 (0.01)     &       0.563 (0.00)      &   0.572 (0.02)  &  \cellcolor{green!=45}0.573 (0.01)  & 0.402 (0.00) &       0.534 (0.01)  & \cellcolor{green!=45}0.553 (0.03)  & 0.544 (0.01)  & 0.389 (0.00)\\
 & MAE & 0.485 (0.06) & \cellcolor{green!=45}0.476 (0.06)  &  0.483 (0.01) &  0.573 (0.01)   & 0.505 (0.00) &  0.497 (0.01) & \cellcolor{green!=45}0.480 (0.00)  & 0.678 (0.00)  &        0.520 (0.00)    &   0.503 (0.07)    &  \cellcolor{green!=45}0.502 (0.02)  & 0.773 (0.00) &    0.56 (0.01) & \cellcolor{green!=45}0.528 (0.05) & 0.541 (0.02)  & 0.813 (0.00) \\
 & MSE & 0.675 (0.01) & \cellcolor{green!=45}0.634 (0.01) & 0.664 (0.02)  & 0.848 (0.04)   &      0.707 (0.01) & 0.683 (0.03)    &  \cellcolor{green!=45}0.655 (0.01)& 1.085 (0.01)   &         0.739 (0.02)      &   0.698 (0.05)    & \cellcolor{green!=45}0.689 (0.02)  &  1.348 (0.01)  &     0.835 (0.04) & \cellcolor{green!=45}0.708 (0.06) & 0.767 (0.01)  & 1.475 (0.01) \\
 & OB1 & 0.938 (0.03) & \cellcolor{green!=45}0.945 (0.03) & 0.939 (0.01) & 0.911 (0.00)    &      0.932 (0.00)  &  0.935 (0.00)    &  \cellcolor{green!=45}0.942 (0.00)& 0.864 (0.00)   &       0.929 (0.00)     &    \cellcolor{green!=45}0.938 (0.03)   & 0.935 (0.01)  &  0.825 (0.00)  &   0.913 (0.00) & \cellcolor{green!=45}0.934 (0.02) & 0.925 (0.01) & 0.807 (0.00) \\
\hline

\end{tabular}
\caption{\small Comparison of techniques for all 3 datasets in full-data and few-shot settings (100, 50, 25, 10\% of data) using BERT-base and GPT2-small as base models. We include one representative loss from each approach -  cross-entropy (CE) baseline (nominal), proposed explicit ordinal loss (MLL) and implicit approaches - Entailment (ENT), Generative (GPT).}
\label{tab:main_table_real}
\addtolength{\tabcolsep}{3.8pt}
\end{minipage}
\end{table*}

Decoder-based text generative models have seen notable advancements in recent years, facilitating the production of coherent and contextually relevant text. The development of models like the GPT series \cite{radford2018improving,noauthororeditor,NEURIPS2020_1457c0d6} has led to state-of-the-art results in text generation and summarization benchmarks.
One of the primary objectives of this paper is to investigate whether these models demonstrate ordinal behavior by accurately capturing the \textbf{inherent order} or ranking of elements in the generated text.

Formally, in the context of OC, for a given a textual input $\bx_i$ which comprises the following words $(w_{i_1},w_{i_2},\dots,w_{i_n})$ and its corresponding ground truth label $L_i$, we append the input and label as $(w_{i_1},w_{i_2},\dots,w_{i_n}, L_i)$.
Next, the parameters $\theta$ of generative model are finetuned s.t. 
\begin{align*}
  \noindent \theta^{*} = \argmax_{\theta} \sum_{i=1}^{N} \log P(w_{i_1},\dots,w_{i_n},L_i; \theta)
\end{align*}
During inference for $i^{th}$ input $\bx_{i}$, the generative model predicts the word $w_{i_{n+1}}$ from the vocabulary $\cv$ which maximizes the conditional probability:
\begin{equation}
   \hat{L} = \argmax_{w_{i_{n+1}} \in \cv}  \{ \log P( w_{i_{n+1}} | w_{i_1},\dots,; \theta^{* }) \}
\end{equation}

 Further, it is possible that the predicted label may not be in the set of ground truth labels i.e. for an $i^{th}$ input $\bx_{i}$ the label $\hat{L}_{i} \notin \bL$ where $\bL$ is the set of $K$ distinct labels (see Appendix~\ref{table:prompt_details_gen}). This is the notorious hallucination problem associated with generative models. 
To mitigate this issue during inference, we compare the conditional log-probabilities of each the $K$ labels
, given the text segment $\bx_{i}$. The class corresponding to the highest probability is proposed as the generated label. That is, given that $\theta^{*}$ represents the learned parameters of LM, the label selection during inference will be s.t.
\begin{equation}
   \hat{L} = \argmax_{L_{j} \in \bL} \log P( L_{j} | w_{i_1},\dots,w_{i_n}; \theta^{*})
\end{equation}
Here $(w_{i_1},w_{i_2},\dots,w_{i_n})$ are the words in input $\bx_i$ and  $(L_{1},L_{2},\dots,L_{K})$ are the labels in the set $\bL$.

For fair comparison with other explicit and implicit approaches, we use \texttt{GPT2-small} as our base model for experiments to maintain similar number of model parameters (with BERT-base). Similar to the entailment approach (\S\ref{sec:implicit_entailment_intro}), we also experiment with informative and un-informative verbalisers here. 

Motivated by recent advancements and the accessibility of open-source Large Language Models (LLMs), and to demonstrate the true potential of the generative approach, we also experiment with Llama-7B \cite{touvron2023llama}, a decoder-based LLM with 7 billion parameters. Pre-trained on trillions of tokens using publicly available data, it achieves state-of-the-art performance, surpassing its larger predecessors like GPT-3 (175B) on the majority of benchmarks.

Note that our approach is different from \texttt{GPT2ForSquenceClassification}\footnote{\url{https://tinyurl.com/am93sjdw}} where the last embedding of the last token is used for classification, which is similar to encoder-model (like BERT) style classification. Instead we train it for a language modelling task
to generate within a fixed set of tokens i.e. the set of labels.

\section{Results and Discussion}
\label{sec:results_discussion}
\begin{figure*}[!hbt]
    \centering

\begin{minipage}{0.23\textwidth}
        \includegraphics[width=\linewidth]{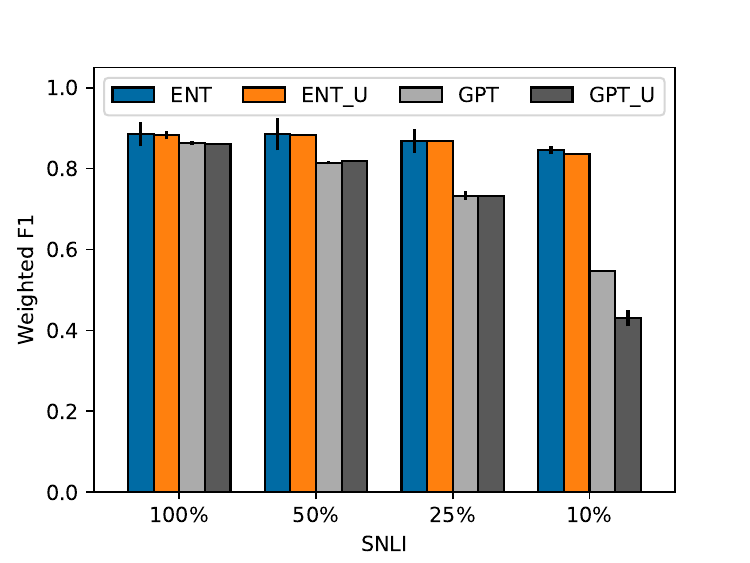}
        \label{fig:plot9}
    \end{minipage}\hfill
    \begin{minipage}{0.23\textwidth}
        \includegraphics[width=\linewidth]{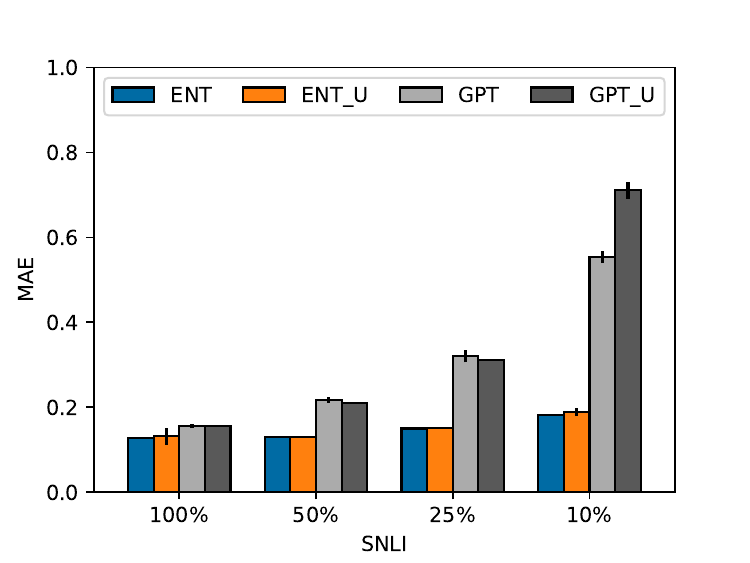}
        \label{fig:plot10}
    \end{minipage}\hfill
    \begin{minipage}{0.23\textwidth}
        \includegraphics[width=\linewidth]{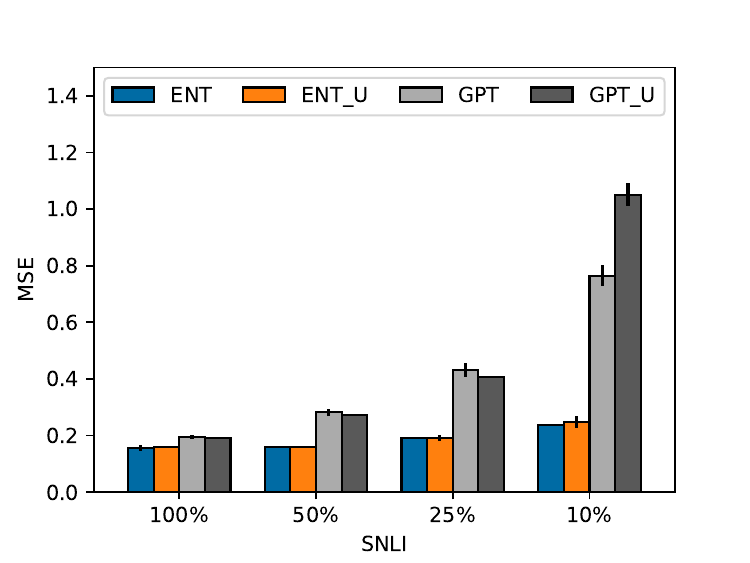}
        \label{fig:plot11}
    \end{minipage}\hfill
    \begin{minipage}{0.23\textwidth}
        \includegraphics[width=\linewidth]{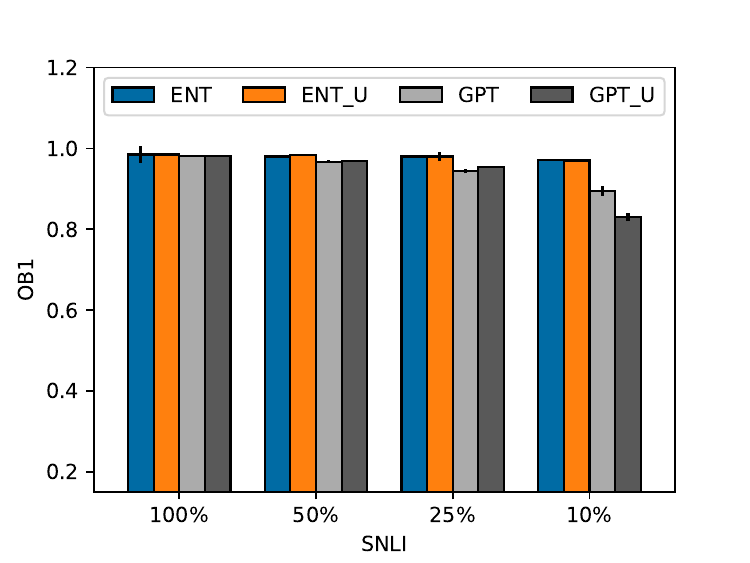}
        \label{fig:plot12}
    \end{minipage}
\vskip -0.65cm
    \begin{minipage}{0.23\textwidth}
        \includegraphics[width=\linewidth]{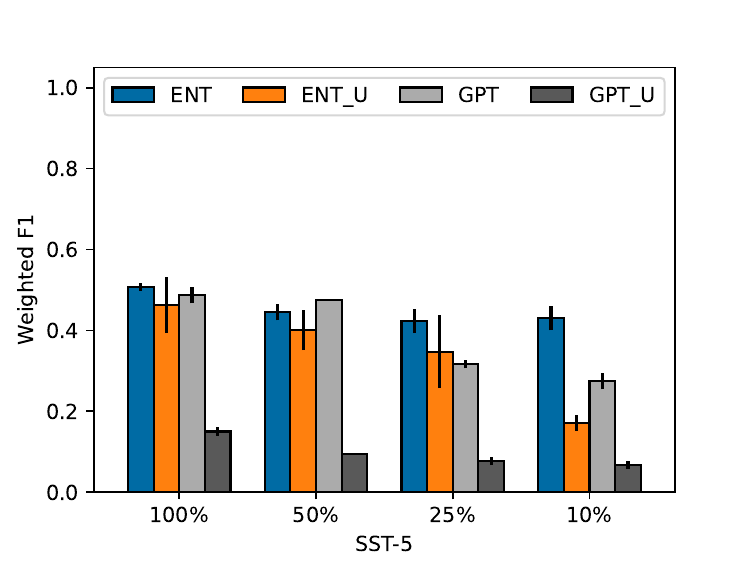}
        \label{fig:plot5}
    \end{minipage}\hfill
    \begin{minipage}{0.23\textwidth}
        \includegraphics[width=\linewidth]{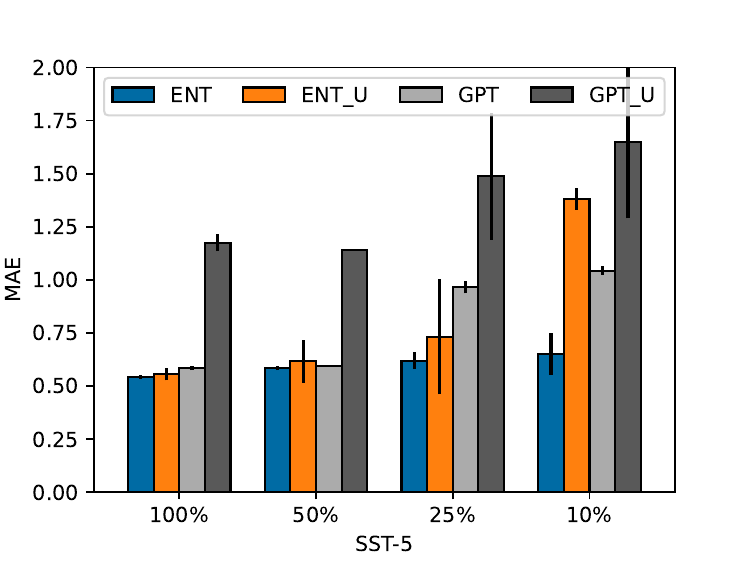}
        \label{fig:plot6}
    \end{minipage}\hfill
    \begin{minipage}{0.23\textwidth}
        \includegraphics[width=\linewidth]{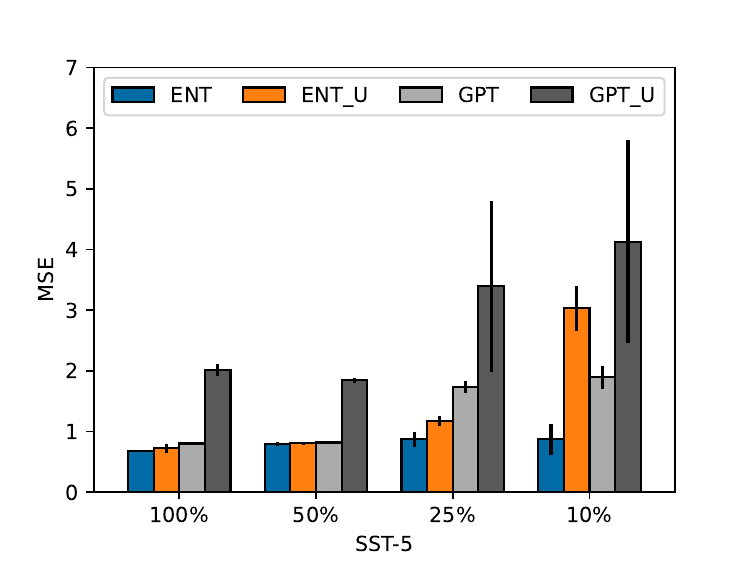}
        \label{fig:plot7}
    \end{minipage}\hfill
    \begin{minipage}{0.23\textwidth}
        \includegraphics[width=\linewidth]{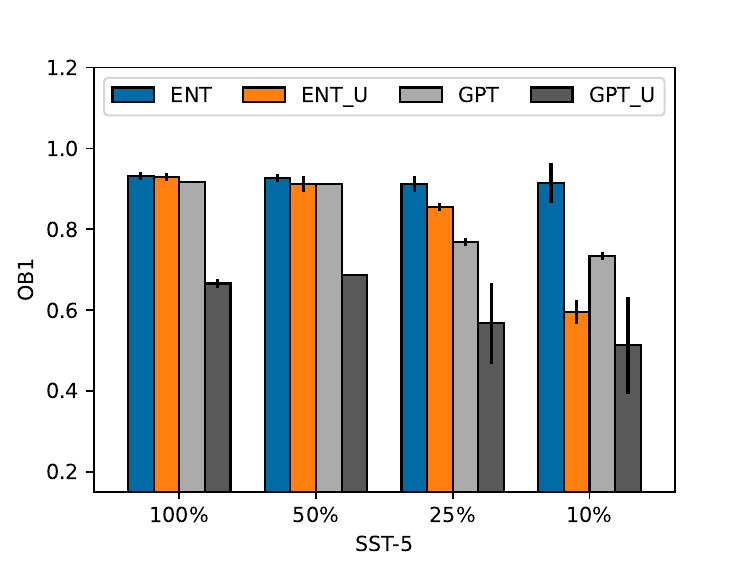}
        \label{fig:plot8}
    \end{minipage}
    \vskip -0.65cm
    \begin{minipage}{0.23\textwidth}
        \includegraphics[width=\linewidth]{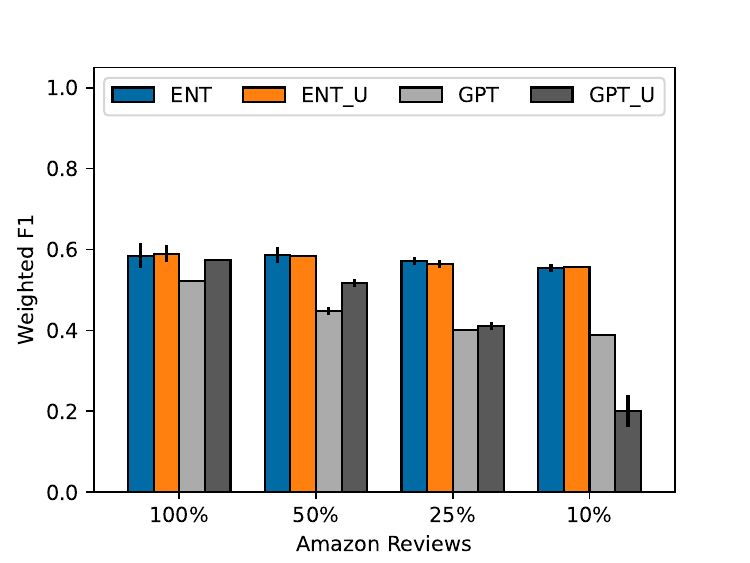}
        \label{fig:plot1}
    \end{minipage}\hfill
    \begin{minipage}{0.23\textwidth}
        \includegraphics[width=\linewidth]{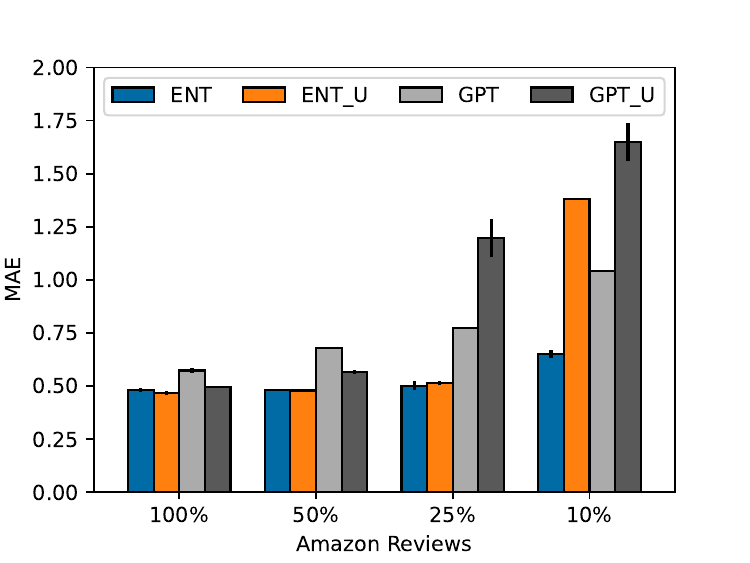}
        \label{fig:plot2}
    \end{minipage}\hfill
    \begin{minipage}{0.23\textwidth}
        \includegraphics[width=\linewidth]{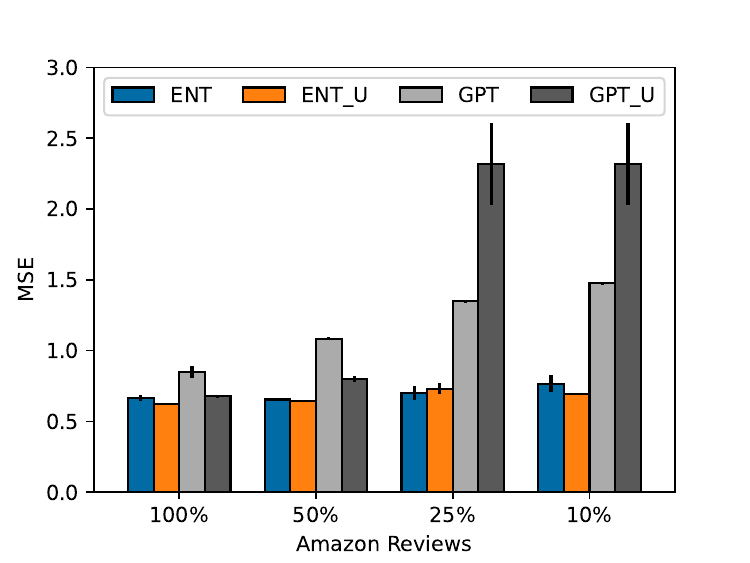}
        \label{fig:plot3}
    \end{minipage}\hfill
    \begin{minipage}{0.23\textwidth}
        \includegraphics[width=\linewidth]{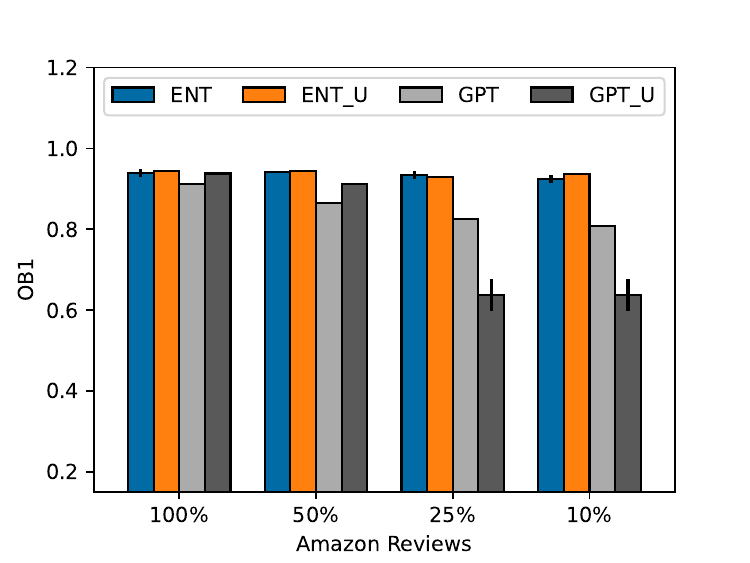}
        \label{fig:plot4}
    \end{minipage}
    
\vskip -0.65cm

    \caption{\small Effect of informative vs uninformative prompts in Implicit approaches on the nominal (F1) and ordinal metrics (MAE, MSE, OB1) on SNLI (\textit{top}), SST5 (\textit{middle}) and Amazon Reviews (\textit{bottom}) datasets. Results averaged across 5 random seeds. (\textit{Notation:} \_U refers to the un-informative verbaliser variant. Weighted-F1 and OB1: $\uparrow$ better, MAE and MSE: $\downarrow$ better.)}
    \label{fig:bar_inf_v_unif}
\end{figure*}

For the explicit loss-based approaches, the experimental results presented in Table \ref{tab:tinybert_main} and \ref{tab:main_table_real} show that our proposed hybrid MLL loss improves the ordinal performance (as measured via MAE, MSE, OB1), compared to CE, without compromising its performance along the nominal dimension (as measured via weighted-F1). A detailed comparison of the performance of MLL against other ordinal loss functions is given in Table~\ref{tab:losses-bert-base} where we further notice that MLL consistently achieves a balanced performance on both ordinal and nominal metrics, compared to other ordinal losses where nominal metrics degrade in general.
But, compared to the results of TinyBERT in Table~\ref{tab:losses-bert-tiny}, the difference in performances is less pronounced \cite{castagnos2022simple}. The observed phenomenon may be explained by the necessity for the smaller base models to rely more on the explicit enforcement of ordinal losses in enhancing performance in OC tasks.  However, as mentioned in \S 2, given that both the losses and metrics are also from the PSR family, and hence are not independent of each other, a bigger base model would improve both the nominal and ordinal metrics simultaneously. Thus, in offline use cases where it is possible to deploy much bigger models, the difference in performance due to various losses is quite minimal. However, in online settings, where it is imperative to deploy lightweight models due to latency reasons, the difference is quite pronounced.

When considering implicit approaches, we observe that the Entailment (ENT) approach performs \textbf{on par} with the MLL loss across all three datasets, without any explicit ordinality enforcing mechanism. It even outperforms CE, MLL, and GPT in almost all data settings for SST-5. This could be attributed to the fact that SST-5 has a significantly lower number of samples ($\sim$12k) compared to SNLI and AR ($\sim$200k). Thus, the ENT approach performs better in few-shot scenarios. Another interesting observation is that for SNLI and AR, although the performances of ENT and MLL are similar, the \textbf{standard deviation} numbers are consistently lower in the case of ENT, making it a reliable and relatively more stable approach to use in low-data settings. This supports our claimed hypothesis that the model is able to leverage natural language label descriptions to inherently enforce ordinality in its predictions.

To examine the impact of incorporating informative verbalisers, we also fine-tuned using un-informative ones, replacing the label descriptions with unrelated words such as "cat"/"lion"/"zebra". Our observations, depicted in Figure~\ref{fig:bar_inf_v_unif}, reveal intriguing findings. When the training data volume is low, the effect of label semantics becomes clearly visible. In the SST-5 dataset (with training data volume $\sim$10k), the \textbf{influence of informative labels} is evident across all settings (refer to \textit{middle} section in Figure~\ref{fig:bar_inf_v_unif}). Similarly, in the AR and SNLI datasets (with training data volume $\sim$250k), the impact of informative labels is noticeable only in the 25\% or 10\% data settings (refer to \textit{top} and \textit{bottom} sections in Figure~\ref{fig:bar_inf_v_unif}). In other cases, the distinction is less clear due to the sheer volume of training data; even un-informative labels yield decent performance, possibly leveraging some spurious correlations, thus rendering the semantics not so effective. This finding also aligns with the hypothesis proposed by \citet{wang2021entailment}, indicating that prompt-based models may not leverage label semantics as expected, with the dependency being influenced by the model and dataset size to some extent. 
On another note, for ENT approach, the inference time is scaled by $\sim$O($K$) ($K$: total labels). But with more carefully engineered verbalisers, similar or even better performance could be achieved even with smaller base models \citep{goodprompt} making it a suitable candidate for ordinal problems.

In our generative (GEN) approach, we initially employed the GPT2-small model for experiments, as it has a comparable number of fine-tunable parameters to the encoder counterpart used in ENT \& Explicit loss approaches. Despite its usual under-performance compared to MLL and ENT, which aligns with the general observation that GPT-2 typically fares worse than BERT-base on most classification tasks \cite{neerudu2023robustness}, we demonstrate its ability to recognize label order by contrasting informative versus un-informative verbalisers, akin to the ENT approach. Similar trends are observed with the Entailment technique, particularly the significant failure of the un-informative verbaliser variant on the SST-5 dataset. Across other datasets, performance in terms of F1-score is comparable to the informative variant, except for 10\% case in SNLI and AR (low-data setting). Notably, for ordinal metrics like MAE and MSE, we observe \textbf{high variance} in AR and SST-5 datasets (Figure~\ref{fig:bar_inf_v_unif}), underscoring the importance of employing informative label verbalizers for stable learning. 
Two common findings emerge from the informative vs un-informative ablations in both ENT and GEN experiments: \textbf{(a)} The hypothesis posited in \cite{webson-pavlick-2022-prompt} holds true for ordinal metrics as well as nominal metrics. \textbf{(b)} Label semantics become significant only in low-data settings.
\begin{table}[H]
\tiny
\begin{tabular}{l|cc|cc|cc}
 & \multicolumn{2}{c}{SST} &  \multicolumn{2}{c}{AR} & \multicolumn{2}{c}{SNLI} \\  \hline
 & I          & U          & I          & U         & I           & U          \\ \hline
F1 &
  \begin{tabular}[c]{@{}c@{}} \textbf{60.845} \\ \textbf{(0.16)}\end{tabular} &
  \begin{tabular}[c]{@{}c@{}}58.55\\ (0.22)\end{tabular} &
  \begin{tabular}[c]{@{}c@{}} \textbf{62.71}\\ \textbf{(0.02)}\end{tabular} &
  \begin{tabular}[c]{@{}c@{}}62.4\\ (0.1)\end{tabular} &
  \begin{tabular}[c]{@{}c@{}}\textbf{90.19}\\ \textbf{(0.15)}\end{tabular} &
  \begin{tabular}[c]{@{}c@{}} 89.08\\ (0.39)\end{tabular} \\
MAE &
  \begin{tabular}[c]{@{}c@{}} \textbf{0.4066}\\ \textbf{(0.003)}\end{tabular} &
  \begin{tabular}[c]{@{}c@{}}0.4327\\ (0.008)\end{tabular} &
  \begin{tabular}[c]{@{}c@{}} \textbf{0.4139}\\ \textbf{(0.001)}\end{tabular} &
  \begin{tabular}[c]{@{}c@{}}0.4186\\ (0.003)\end{tabular} &
  \begin{tabular}[c]{@{}c@{}}\textbf{0.1068}\\ \textbf{(0.001)}\end{tabular} &
  \begin{tabular}[c]{@{}c@{}} 0.1188\\ (0.004)\end{tabular} \\
MSE &
  \begin{tabular}[c]{@{}c@{}} \textbf{0.4538} \\ \textbf{(0.007)}\end{tabular} &
  \begin{tabular}[c]{@{}c@{}}0.4918\\ (0.01)\end{tabular} &
  \begin{tabular}[c]{@{}c@{}} \textbf{0.5087}\\ \textbf{(0.005)}\end{tabular} &
  \begin{tabular}[c]{@{}c@{}}0.5182\\ (0.008)\end{tabular} &
  \begin{tabular}[c]{@{}c@{}}\textbf{0.1241}\\ \textbf{(0.001)}\end{tabular} &
  \begin{tabular}[c]{@{}c@{}} 0.1366\\ (0.013)\end{tabular} \\
OB1 &
  \begin{tabular}[c]{@{}c@{}} \textbf{0.9786}\\ \textbf{(0.001)}\end{tabular} &
  \begin{tabular}[c]{@{}c@{}}0.9750\\ (0.003)\end{tabular} &
  \begin{tabular}[c]{@{}c@{}} \textbf{0.963}\\ \textbf{(0.001)}\end{tabular} &
  \begin{tabular}[c]{@{}c@{}}0.9595\\ (0.003)\end{tabular} &
  \begin{tabular}[c]{@{}c@{}}\textbf{0.9913}\\ \textbf{(0.001)}\end{tabular} &
  \begin{tabular}[c]{@{}c@{}} 0.9911\\ (0.002)\end{tabular}
\end{tabular}
\caption{\small Effect of informative (I) vs un-informative (U) verbalisers on Llama-7B}
\label{tab:llama}
\end{table}
    \vspace{-0.35cm}
    \begin{figure}[h]
    \centering
    \includegraphics[width=0.35\textwidth]{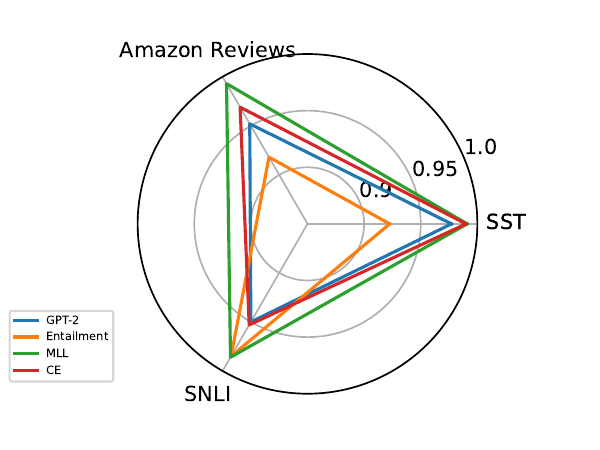}
        \vspace{-0.25cm}
    \caption{\small \% samples following UM property.}
    \label{fig:unimodal}
\end{figure}

The purpose of presenting GPT-2 results was to illustrate the true potential of generative approaches, through LLMs, as introduced in \S\ref{sec:implicit_decoder_intro}.
We leverage the \textit{Llama-Adapter} technique \cite{adapter} which introduces only 1.2M tunable parameters over the base model by prepending a set of learnable adaption verbalisers to the word tokens at higher transformer layers and show that it still outperforms all explicit \& implicit approaches presented above (refer Table~\ref{tab:llama}). 
While we acknowledge that the enhanced ordinal performance may be attributed to the increased classification performance resulting from \textbf{larger model size}, we refer to Table~\ref{tab:llama} to demonstrate that even with larger base models, the distinction between informative and uninformative verbalisers persists. This underscores the model's ability to recognize label order.
In Figure~\ref{fig:unimodal} we show the \% of samples which follow the unimodality (UM) property empirically. Although there is no theoretical guarantee for both our proposed explicit \& implicit strategies, we observe \textbf{$\sim 85$ \% samples} satisfy this property which justifies the increased ordinal performance of the above discussed approaches.

\section{Conclusion}
\label{sec:conclusion}

\begin{table}[ht]
    \centering
    \small
    \begin{tabular}{c|c}
        Technique & Recommended Setting \\ \hline
        MLL & high-data regime, high \% unimodality \\
        ENT &  low-data regime, inference time $\sim O(K)$\\
        LLM & high-data regime, compute-intensive, \\
        & hallucinates in low-data setting \\
    \end{tabular}
    \caption{\small Recommended techniques for various settings}
    \label{tab:recommendations}
\end{table}

This paper presents an unified analysis of explicit and implicit strategies for addressing OC. It is the first study to thoroughly examine and compare these approaches from both theoretical and empirical standpoints. Our analysis (summarized in Table~\ref{tab:recommendations}) reveals that MLL demonstrates balanced performance across ordinal and nominal metrics, unlike existing explicit losses. However, in few-shot scenarios, ENT is preferred for its ability to achieve optimal performance with fewer examples, leveraging label semantics. Furthermore, we highlight the importance of providing informative verbalisers in low-data settings, resulting in reduced variance and improved outcomes. However, the distinction between strategies becomes less clear with increasing data. In full-data scenarios, fine-tuning Llama-7B-Adapter surpasses previous approaches due to its substantial model size. Interestingly, even with such a large base model, the impact of adding informative verbalisers remains apparent, indicating its recognition of label order.
We hope that our work will serve as a benchmark encompassing multitude of approaches, providing a foundation for future efforts to address OC in NLP.

\section{Limitations}
In this study,
we don't consider the effect of calibration techniques on the explicit approaches \citep{NEURIPS2019_8ca01ea9}, as the techniques employed are largely identical to those used in nominal classification, offering no distinct or novel methodologies specifically for OC tasks.
Also for implicit approaches, a more deeper analysis is required on how implicit methods like LLMs \& PLMs implicitly capture ordinality. It's not always analogous to how humans use task instructions as shown in \citet{webson-pavlick-2022-prompt}. Furthermore, in this study we limit ourselves to only finetuning-based OC approaches. However, it would also be interesting to explore OC through the lens of in-context learning (ICL) for generative approaches. Also for the generative approach, we make the assumption that the label word will not break further into multiple tokens by re-mapping original labels to simpler words (see Appendix~\ref{table:prompt_details_gen}). This avoids having to account for multiple token probabilities when taking the argmax. Without this some sort of normalization would be required across the entire generation length to compare different outputs. We leave these discussions for future work. Even though we notice that LLMs such as Llama-7B outperform all the other models in full data settings, there are certain challenges in terms of compute resources and inference time. Additionally, finetuning Llama-7B is susceptible to hallucinations in low-data settings \cite{pmlr-v139-zhao21c}, which is why we don't report LLM results for the few-shot case.

\bibliography{custom}

\begin{thebibliography}{37}
\expandafter\ifx\csname natexlab\endcsname\relax\def\natexlab#1{#1}\fi

\bibitem[{Beckham and Pal(2017)}]{beckham2017unimodal}
Christopher Beckham and Christopher Pal. 2017.
\newblock Unimodal probability distributions for deep ordinal classification.
\newblock In \emph{International Conference on Machine Learning}, pages 411--419. PMLR.

\bibitem[{Bowman et~al.(2015)Bowman, Angeli, Potts, and Manning}]{bowman-etal-2015-large}
Samuel~R. Bowman, Gabor Angeli, Christopher Potts, and Christopher~D. Manning. 2015.
\newblock \href {https://doi.org/10.18653/v1/D15-1075} {A large annotated corpus for learning natural language inference}.
\newblock In \emph{Proceedings of the 2015 Conference on Empirical Methods in Natural Language Processing}, pages 632--642, Lisbon, Portugal. Association for Computational Linguistics.

\bibitem[{Boyd and Vandenberghe(2004)}]{boyd2004convex}
Stephen~P Boyd and Lieven Vandenberghe. 2004.
\newblock \emph{Convex optimization}.
\newblock Cambridge university press.

\bibitem[{Brown et~al.(2020)Brown, Mann, Ryder, Subbiah, Kaplan, Dhariwal, Neelakantan, Shyam, Sastry, Askell, Agarwal, Herbert-Voss, Krueger, Henighan, Child, Ramesh, Ziegler, Wu, Winter, Hesse, Chen, Sigler, Litwin, Gray, Chess, Clark, Berner, McCandlish, Radford, Sutskever, and Amodei}]{NEURIPS2020_1457c0d6}
Tom Brown, Benjamin Mann, Nick Ryder, Melanie Subbiah, Jared~D Kaplan, Prafulla Dhariwal, Arvind Neelakantan, Pranav Shyam, Girish Sastry, Amanda Askell, Sandhini Agarwal, Ariel Herbert-Voss, Gretchen Krueger, Tom Henighan, Rewon Child, Aditya Ramesh, Daniel Ziegler, Jeffrey Wu, Clemens Winter, Chris Hesse, Mark Chen, Eric Sigler, Mateusz Litwin, Scott Gray, Benjamin Chess, Jack Clark, Christopher Berner, Sam McCandlish, Alec Radford, Ilya Sutskever, and Dario Amodei. 2020.
\newblock \href {https://proceedings.neurips.cc/paper_files/paper/2020/file/1457c0d6bfcb4967418bfb8ac142f64a-Paper.pdf} {Language models are few-shot learners}.
\newblock In \emph{Advances in Neural Information Processing Systems}, volume~33, pages 1877--1901. Curran Associates, Inc.

\bibitem[{Cao et~al.(2020)Cao, Mirjalili, and Raschka}]{Cao_2020}
Wenzhi Cao, Vahid Mirjalili, and Sebastian Raschka. 2020.
\newblock \href {https://doi.org/10.1016/j.patrec.2020.11.008} {Rank consistent ordinal regression for neural networks with application to age estimation}.
\newblock \emph{Pattern Recognition Letters}, 140:325--331.

\bibitem[{Castagnos et~al.(2022)Castagnos, Mihelich, and Dognin}]{castagnos2022simple}
Fran{\c{c}}ois Castagnos, Martin Mihelich, and Charles Dognin. 2022.
\newblock A simple log-based loss function for ordinal text classification.
\newblock In \emph{Proceedings of the 29th International Conference on Computational Linguistics}, pages 4604--4609.

\bibitem[{da~Costa et~al.(2008)da~Costa, Alonso, and Cardoso}]{da2008unimodal}
Joaquim F~Pinto da~Costa, Hugo Alonso, and Jaime~S Cardoso. 2008.
\newblock The unimodal model for the classification of ordinal data.
\newblock \emph{Neural Networks}, 21(1):78--91.

\bibitem[{Dang et~al.(2020)Dang, Moreno-Garc{\'\i}a, and De~la Prieta}]{dang2020sentiment}
Nhan~Cach Dang, Mar{\'\i}a~N Moreno-Garc{\'\i}a, and Fernando De~la Prieta. 2020.
\newblock Sentiment analysis based on deep learning: A comparative study.
\newblock \emph{Electronics}, 9(3):483.

\bibitem[{de~la Torre et~al.(2017)de~la Torre, Puig, and Valls}]{article}
Jordi de~la Torre, Domenec Puig, and Aida Valls. 2017.
\newblock \href {https://doi.org/10.1016/j.patrec.2017.05.018} {Weighted kappa loss function for multi-class classification of ordinal data in deep learning}.
\newblock \emph{Pattern Recognition Letters}.

\bibitem[{Devlin et~al.(2019)Devlin, Chang, Lee, and Toutanova}]{devlin-etal-2019-bert}
Jacob Devlin, Ming-Wei Chang, Kenton Lee, and Kristina Toutanova. 2019.
\newblock \href {https://doi.org/10.18653/v1/N19-1423} {{BERT}: Pre-training of deep bidirectional transformers for language understanding}.
\newblock In \emph{Proceedings of the 2019 Conference of the North {A}merican Chapter of the Association for Computational Linguistics: Human Language Technologies, Volume 1 (Long and Short Papers)}, pages 4171--4186, Minneapolis, Minnesota. Association for Computational Linguistics.

\bibitem[{Du et~al.(2019)Du, Lee, Li, Wang, and Zhai}]{du2019gradient}
Simon Du, Jason Lee, Haochuan Li, Liwei Wang, and Xiyu Zhai. 2019.
\newblock Gradient descent finds global minima of deep neural networks.
\newblock In \emph{International conference on machine learning}, pages 1675--1685. PMLR.

\bibitem[{Díaz and Marathe(2019)}]{diaz2019soft}
Raúl Díaz and Amit Marathe. 2019.
\newblock \href {https://doi.org/10.1109/CVPR.2019.00487} {Soft labels for ordinal regression}.
\newblock In \emph{2019 IEEE/CVF Conference on Computer Vision and Pattern Recognition (CVPR)}, pages 4733--4742.

\bibitem[{Gneiting and Raftery(2007)}]{gneiting2007strictly}
Tilmann Gneiting and Adrian~E Raftery. 2007.
\newblock Strictly proper scoring rules, prediction, and estimation.
\newblock \emph{Journal of the American statistical Association}, 102(477):359--378.

\bibitem[{Iannario and Piccolo(2011)}]{iannario2011cub}
Maria Iannario and Domenico Piccolo. 2011.
\newblock Cub models: Statistical methods and empirical evidence.
\newblock \emph{Modern Analysis of Customer Surveys: with applications using R}, pages 231--258.

\bibitem[{Jiao et~al.(2020)Jiao, Yin, Shang, Jiang, Chen, Li, Wang, and Liu}]{jiao-etal-2020-tinybert}
Xiaoqi Jiao, Yichun Yin, Lifeng Shang, Xin Jiang, Xiao Chen, Linlin Li, Fang Wang, and Qun Liu. 2020.
\newblock \href {https://doi.org/10.18653/v1/2020.findings-emnlp.372} {{T}iny{BERT}: Distilling {BERT} for natural language understanding}.
\newblock In \emph{Findings of the Association for Computational Linguistics: EMNLP 2020}, pages 4163--4174, Online. Association for Computational Linguistics.

\bibitem[{Jin et~al.(2022)Jin, Cheng, Shen, Chen, and Ren}]{goodprompt}
Woojeong Jin, Yu~Cheng, Yelong Shen, Weizhu Chen, and Xiang Ren. 2022.
\newblock \href {https://doi.org/10.18653/v1/2022.acl-long.197} {A good prompt is worth millions of parameters: Low-resource prompt-based learning for vision-language models}.
\newblock In \emph{Proceedings of the 60th Annual Meeting of the Association for Computational Linguistics (Volume 1: Long Papers)}, pages 2763--2775, Dublin, Ireland. Association for Computational Linguistics.

\bibitem[{Kawaguchi et~al.(2019)Kawaguchi, Huang, and Kaelbling}]{kawaguchi2019every}
Kenji Kawaguchi, Jiaoyang Huang, and Leslie~Pack Kaelbling. 2019.
\newblock Every local minimum value is the global minimum value of induced model in nonconvex machine learning.
\newblock \emph{Neural Computation}, 31(12):2293--2323.

\bibitem[{Keung et~al.(2020)Keung, Lu, Szarvas, and Smith}]{keung-etal-2020-multilingual}
Phillip Keung, Yichao Lu, Gy{\"o}rgy Szarvas, and Noah~A. Smith. 2020.
\newblock \href {https://doi.org/10.18653/v1/2020.emnlp-main.369} {The multilingual {A}mazon reviews corpus}.
\newblock In \emph{Proceedings of the 2020 Conference on Empirical Methods in Natural Language Processing (EMNLP)}, pages 4563--4568, Online. Association for Computational Linguistics.

\bibitem[{Kull et~al.(2019)Kull, Perello~Nieto, K\"{a}ngsepp, Silva~Filho, Song, and Flach}]{NEURIPS2019_8ca01ea9}
Meelis Kull, Miquel Perello~Nieto, Markus K\"{a}ngsepp, Telmo Silva~Filho, Hao Song, and Peter Flach. 2019.
\newblock \href {https://proceedings.neurips.cc/paper_files/paper/2019/file/8ca01ea920679a0fe3728441494041b9-Paper.pdf} {Beyond temperature scaling: Obtaining well-calibrated multi-class probabilities with dirichlet calibration}.
\newblock In \emph{Advances in Neural Information Processing Systems}, volume~32. Curran Associates, Inc.

\bibitem[{Lakshminarayanan et~al.(2017)Lakshminarayanan, Pritzel, and Blundell}]{lakshminarayanan2017simple}
Balaji Lakshminarayanan, Alexander Pritzel, and Charles Blundell. 2017.
\newblock Simple and scalable predictive uncertainty estimation using deep ensembles.
\newblock \emph{Advances in neural information processing systems}, 30.

\bibitem[{Liu(2020)}]{liu2020yelp}
Zefang Liu. 2020.
\newblock \href {http://arxiv.org/abs/2012.06690} {Yelp review rating prediction: Machine learning and deep learning models}.

\bibitem[{M{\'e}rigot et~al.(2021)M{\'e}rigot, Santambrogio, and Sarrazin}]{merigot2021non}
Quentin M{\'e}rigot, Filippo Santambrogio, and Cl{\'e}ment Sarrazin. 2021.
\newblock Non-asymptotic convergence bounds for wasserstein approximation using point clouds.
\newblock \emph{Advances in Neural Information Processing Systems}, 34:12810--12821.

\bibitem[{Merkle and Steyvers(2013)}]{merkle2013choosing}
Edgar~C Merkle and Mark Steyvers. 2013.
\newblock Choosing a strictly proper scoring rule.
\newblock \emph{Decision Analysis}, 10(4):292--304.

\bibitem[{Neerudu et~al.(2023)Neerudu, Oota, Marreddy, Kagita, and Gupta}]{neerudu2023robustness}
Pavan Kalyan~Reddy Neerudu, Subba~Reddy Oota, Mounika Marreddy, Venkateswara~Rao Kagita, and Manish Gupta. 2023.
\newblock On robustness of finetuned transformer-based nlp models.
\newblock \emph{arXiv preprint arXiv:2305.14453}.

\bibitem[{Pilanci and Ergen(2020)}]{pilanci2020neural}
Mert Pilanci and Tolga Ergen. 2020.
\newblock Neural networks are convex regularizers: Exact polynomial-time convex optimization formulations for two-layer networks.
\newblock In \emph{International Conference on Machine Learning}, pages 7695--7705. PMLR.

\bibitem[{Radford et~al.(2018{\natexlab{a}})Radford, Narasimhan, Salimans, and Sutskever}]{radford2018improving}
Alec Radford, Karthik Narasimhan, Tim Salimans, and Ilya Sutskever. 2018{\natexlab{a}}.
\newblock Improving language understanding by generative pre-training.
\newblock \emph{OpenAI}.

\bibitem[{Radford et~al.(2018{\natexlab{b}})Radford, Wu, Child, Luan, Amodei, and Sutskever}]{noauthororeditor}
Alec Radford, Jeffrey Wu, Rewon Child, David Luan, Dario Amodei, and Ilya Sutskever. 2018{\natexlab{b}}.
\newblock \href {https://d4mucfpksywv.cloudfront.net/better-language-models/language-models.pdf} {Language models are unsupervised multitask learners}.

\bibitem[{Rubner et~al.(2000)Rubner, Tomasi, and Guibas}]{rubner2000earth}
Yossi Rubner, Carlo Tomasi, and Leonidas~J Guibas. 2000.
\newblock The earth mover's distance as a metric for image retrieval.
\newblock \emph{International journal of computer vision}, 40(2):99.

\bibitem[{S{\'a}nchez-Hevia et~al.(2022)S{\'a}nchez-Hevia, Gil-Pita, Utrilla-Manso, and Rosa-Zurera}]{sanchez2022age}
H{\'e}ctor~A S{\'a}nchez-Hevia, Roberto Gil-Pita, Manuel Utrilla-Manso, and Manuel Rosa-Zurera. 2022.
\newblock Age group classification and gender recognition from speech with temporal convolutional neural networks.
\newblock \emph{Multimedia Tools and Applications}, 81(3):3535--3552.

\bibitem[{Socher et~al.(2013)Socher, Perelygin, Wu, Chuang, Manning, Ng, and Potts}]{socher-etal-2013-recursive}
Richard Socher, Alex Perelygin, Jean Wu, Jason Chuang, Christopher~D. Manning, Andrew Ng, and Christopher Potts. 2013.
\newblock \href {https://aclanthology.org/D13-1170} {Recursive deep models for semantic compositionality over a sentiment treebank}.
\newblock In \emph{Proceedings of the 2013 Conference on Empirical Methods in Natural Language Processing}, pages 1631--1642, Seattle, Washington, USA. Association for Computational Linguistics.

\bibitem[{Touvron et~al.(2023)Touvron, Lavril, Izacard, Martinet, Lachaux, Lacroix, Rozière, Goyal, Hambro, Azhar, Rodriguez, Joulin, Grave, and Lample}]{touvron2023llama}
Hugo Touvron, Thibaut Lavril, Gautier Izacard, Xavier Martinet, Marie-Anne Lachaux, Timothée Lacroix, Baptiste Rozière, Naman Goyal, Eric Hambro, Faisal Azhar, Aurelien Rodriguez, Armand Joulin, Edouard Grave, and Guillaume Lample. 2023.
\newblock \href {http://arxiv.org/abs/2302.13971} {Llama: Open and efficient foundation language models}.
\newblock Cite arxiv:2302.13971.

\bibitem[{Wang et~al.(2021)Wang, Fang, Khabsa, Mao, and Ma}]{wang2021entailment}
Sinong Wang, Han Fang, Madian Khabsa, Hanzi Mao, and Hao Ma. 2021.
\newblock \href {http://arxiv.org/abs/2104.14690} {Entailment as few-shot learner}.

\bibitem[{Webson and Pavlick(2022)}]{webson-pavlick-2022-prompt}
Albert Webson and Ellie Pavlick. 2022.
\newblock \href {https://doi.org/10.18653/v1/2022.naacl-main.167} {Do prompt-based models really understand the meaning of their prompts?}
\newblock In \emph{Proceedings of the 2022 Conference of the North American Chapter of the Association for Computational Linguistics: Human Language Technologies}, pages 2300--2344, Seattle, United States. Association for Computational Linguistics.

\bibitem[{Wojtowytsch(2023)}]{wojtowytsch2023stochastic}
Stephan Wojtowytsch. 2023.
\newblock Stochastic gradient descent with noise of machine learning type part i: Discrete time analysis.
\newblock \emph{Journal of Nonlinear Science}, 33(3):45.

\bibitem[{Yamasaki(2022)}]{yamasaki2022unimodal}
Ryoya Yamasaki. 2022.
\newblock \href {https://openreview.net/forum?id=1l0sClLiPc} {Unimodal likelihood models for ordinal data}.
\newblock \emph{Transactions on Machine Learning Research}.

\bibitem[{Zhang et~al.(2023)Zhang, Han, Zhou, Hu, Yan, Lu, Li, Gao, and Qiao}]{adapter}
Renrui Zhang, Jiaming Han, Aojun Zhou, Xiangfei Hu, Shilin Yan, Pan Lu, Hongsheng Li, Peng Gao, and Yu~Qiao. 2023.
\newblock Llama-adapter: Efficient fine-tuning of language models with zero-init attention.
\newblock \emph{arXiv preprint arXiv:2303.16199}.

\bibitem[{Zhao et~al.(2021)Zhao, Wallace, Feng, Klein, and Singh}]{pmlr-v139-zhao21c}
Zihao Zhao, Eric Wallace, Shi Feng, Dan Klein, and Sameer Singh. 2021.
\newblock \href {https://proceedings.mlr.press/v139/zhao21c.html} {Calibrate before use: Improving few-shot performance of language models}.
\newblock In \emph{Proceedings of the 38th International Conference on Machine Learning}, volume 139 of \emph{Proceedings of Machine Learning Research}, pages 12697--12706. PMLR.

\end{thebibliography}
\clearpage

\appendix

\section{Comparison of different loss based approaches}
\label{app:loss_comparison}

We follow \citet{castagnos2022simple} and using \textit{TinyBERT} as the backbone mode, ran CE, OLL, MLL, SOFT, EMD, CORAL, WKL and VS-SL on the four real datasets mentioned in appendix \ref{app:datasets} and compute the both the nominal and ordinal metrics mentioned in appendix \ref{app:metrics}. The results are given in Table \ref{tab:losses-bert-tiny}.

\begin{table}[!ht]
\tiny
    \centering
    \begin{tabular}{p{0.75cm}|p{0.5cm}p{0.75cm}p{0.25cm}p{0.25cm}p{0.25cm}p{0.25cm}}
        \textbf{Dataset} & \textbf{Loss } & \textbf{LR} & \textbf{F1} & \textbf{MSE} & 
        \textbf{MAE} & \textbf{OB1} \\ \hline 
        \textbf{SST5} & CE & 0.000025 & 0.357 (0.01) & 1.197 (0.00) & 0.768 (0.02) & 0.852 (0.00) \\ 
        \textbf{} & MLL & 0.0001 & \textbf{0.378 (0.02)} & 1.125 (0.01) & 0.742 (0.01) & 0.863 (0.00) \\ 
        \textbf{} & OLL & 0.000025 & 0.359 (0.02) & \textbf{1.055 (0.00)} & \textbf{0.740 (0.01)} & \textbf{0.870 (0.00)} \\ 
        \textbf{} & WKL & 0.000075 & 0.366 (0.00) & 1.250 (0.00)  & 0.809 (0.02) & 0.847 (0.01) \\ 
        \textbf{} & SOFT & 0.0001 & 0.382 (0.00) & 1.152 (0.00) & 0.751 (0.03) & 0.856 (0.00) \\ 
        \textbf{} & EMD & 0.0001 & 0.354 (0.02) & 1.125 (0.03) & 0.744 (0.02) & 0.857 (0.03) \\ 
        \textbf{} & CORAL & 0.0001 & 0.109 (0.00) & 2.739 (0.01) & 1.281 (0.06) & 0.641 (0.00) \\ 
        \textbf{} & VS_SL & 0.000025 & 0.233 (0.05) & 2.099 (0.02) & 1.087 (0.09) & 0.710 (0.01)\\ \\
        \textbf{Amazon reviews} & CE & 0.0001 & 0.543 (0.02) & 0.904 (0.04) & 0.581 (0.01) & 0.903 (0.00) \\ 
        \textbf{} & MLL & 0.000075 & \textbf{0.544 (0.00)} & 0.819 (0.01) & \textbf{0.568 (0.00)} & 0.915 (0.00) \\ 
        \textbf{} & OLL & 0.0001 & 0.530 (0.00) & \textbf{0.788 (0.00)} & 0.571 (0.00) & \textbf{0.924 (0.00)} \\ 
        \textbf{} & WKL & 0.00005 & 0.515 (0.00) & 0.871 (0.00) & 0.594 (0.01) & 0.907 (0.05) \\ 
        \textbf{} & SOFT & 0.000075 & 0.537 (0.00) & 0.904 (0.00) & 0.586 (0.01) & 0.903 (0.00) \\ 
        \textbf{} & EMD & 0.00005 & 0.534 (0.00) & 0.885 (0.00) & 0.584 (0.00) & 0.904 (0.00) \\ 
        \textbf{} & CORAL & 0.0001 & 0.349 (0.00) & 1.363 (0.01) & 0.781 (0.01) & 0.890 (0.00) \\ 
        \textbf{} & VS_SL & 0.00005 & 0.377 (0.05) & 1.535 (0.08)  &  0.853 (0.03) & 0.793 (0.02) \\ \\
        \textbf{SNLI} & CE & 0.00005 & 0.821 (0.00) & 0.264 (0.00) & 0.208 (0.02) & 0.972 (0.01) \\ 
        \textbf{} & MLL & 0.000075 & \textbf{0.832 (0.00)} & 0.257 (0.01) & \textbf{0.205 (0.01)} & 0.974 (0.00) \\ 
        \textbf{} & OLL & 0.0001 & 0.803 (0.01) & \textbf{0.250 (0.03)} & 0.217 (0.01) & \textbf{0.980 (0.02)} \\ 
        \textbf{} & WKL & 0.0001 & 0.782 (0.01) & 0.289 (0.03) & 0.242 (0.02) & 0.975 (0.00) \\ 
        \textbf{} & SOFT & 0.0001 & 0.824 (0.00) & 0.257 (0.01) & 0.203 (0.00) & 0.972 (0.00) \\ 
        \textbf{} & EMD & 0.0001 & 0.826 (0.00) & 0.250 (0.00) & 0.199 (0.01) & 0.975 (0.00) \\ 
        \textbf{} & CORAL & 0.0001 & 0.815 (0.00) & 0.251 (0.02) & 0.207 (0.02) & 0.977 (0.01) \\ 
        \textbf{} & VS_SL & 0.00005	& 0.778 (0.02) & 0.339 (0.04)  & 0.261 (0.03) & 0.955 (0.00)\\
    \end{tabular}
    \caption{Loss functions comparison on three datasets using BERT-tiny architecture}
    \label{tab:losses-bert-tiny}
\end{table}

In section \ref{sec:train_detail}, in order to facilitate fair comparison across implicit and explicit approaches, we repeat the same experiments with \textit{BERT-base-uncased}. The results are given in Table \ref{tab:losses-bert-base}

\begin{table}[!ht]
\tiny
    \centering
    \begin{tabular}{p{0.75cm}|p{0.5cm}p{1.0cm}p{0.25cm}p{0.25cm}p{0.25cm}p{0.25cm}}
        \textbf{Dataset} & \textbf{Loss } & \textbf{LR} & \textbf{F1} & \textbf{MSE} & \textbf{MAE} & \textbf{OB1} \\ \hline
        \textbf{SST5} & CE & 1.00E-05 & 0.484 (0.01) & 0.761 (0.04) & 0.576 (0.02) & 0.925 (0.01) \\ 
        \textbf{} & MLL & 5.00E-05 & \textbf{0.492 (0.01)} & 0.757 (0.01) & \textbf{0.575 (0.02)} & \textbf{0.931 (0.00)} \\ 
        \textbf{} & OLL & 5.00E-05 & 0.456 (0.01) & \textbf{0.735 (0.00)} & 0.586 (0.01) & 0.927 (0.00)\\ 
        \textbf{} & WKL & 2.50E-05 & 0.488 (0.00) & 0.751 (0.01) & 0.584 (0.03) & 0.926 (0.03) \\ 
        \textbf{} & SOFT & 1.00E-05 & 0.486 (0.00) & 0.761 (0.01) & 0.581 (0.02) & 0.924 (0.01) \\ 
        \textbf{} & EMD & 7.50E-05 & 0.466 (0.01) & 0.770 (0.02) & 0.592 (0.01) & 0.916 (0.00) \\ 
        \textbf{} & CORAL & 1.00E-05 & 0.450 (0.02) & 0.927 (0.00) & 0.717 (0.01) & 0.923 (0.00) \\ 
        \textbf{} & VS_SL & 1.00E-05 & 0.2764 (0.04) & 2.663 (0.02) & 1.204 (0.08) & 0.670 (0.08) \\ \\
        \textbf{Amazon reviews} & CE & 1.00E-05 & 0.586 (0.04) & 0.675 (0.01) & 0.485 (0.06) & 0.938 (0.03) \\ 
        \textbf{} & MLL & 5.00E-05 & \textbf{0.589 (0.04)} & 0.634 (0.01) & \textbf{0.476 (0.06)} & 0.945 (0.03) \\ 
        \textbf{} & OLL & 2.50E-05 & 0.586 (0.00) & \textbf{0.622 (0.00)} & 0.477 (0.00) & \textbf{0.948 (0.00)} \\ 
        \textbf{} & WKL & 1.00E-05 & 0.582 (0.02) & 0.641 (0.03)& 0.482 (0.07) & 0.944 (0.01) \\ 
        \textbf{} & SOFT & 1.00E-05 & 0.584 (0.00) & 0.681 (0.00) & 0.489 (0.00) & 0.937 (0.00) \\ 
        \textbf{} & EMD & 5.00E-05 & 0.580 (0.00) & 0.652 (0.01) & 0.490 (0.00) & 0.942 (0.00) \\ 
        \textbf{} & CORAL & 5.00E-05 & 0.406 (0.00) & 1.231 (0.01) & 0.865 (0.01) & 0.966 (0.00) \\ 
        \textbf{} & VS_SL & 7.50E-05 & 0.328 (0.01) & 1.613 (0.02) & 0.907 (0.05) & 0.786 (0.00) \\ \\
        \textbf{SNLI} & CE & 5.00E-05 & 0.890 (0.02) & 0.152 (0.04) & 0.123 (0.01) & 0.985 (0.01) \\ 
        \textbf{} & MLL & 5.00E-05 & \textbf{0.891 (0.02)} & 0.149 (0.04) & 0.122 (0.02) & 0.986 (0.01) \\ 
        \textbf{} & OLL & 5.00E-05 & 0.890 (0.01) & \textbf{0.143 (0.00)} & \textbf{0.121 (0.01)} & \textbf{0.989 (0.00)} \\ 
        \textbf{} & WKL & 1.00E-05 & 0.865 (0.01) & 0.184 (0.01) & 0.152 (0.00) & 0.984 (0.00) \\ 
        \textbf{} & SOFT & 5.00E-05 & 0.889 (0.02) & 0.160 (0.02) & 0.127 (0.04) & 0.980 (0.00) \\ 
        \textbf{} & EMD & 1.00E-05 & 0.890 (0.00) & 0.154 (0.04) & 0.126 (0.02) & 0.985 (0.00) \\ 
        \textbf{} & CORAL & 7.50E-05 & 0.885 (0.00) & 0.189 (0.01) & 0.128 (0.01) & 0.988 (0.00) \\ 
        \textbf{} & VS_SL & 2.5E-05 & 0.810 (0.01) & 0.300 (0.01) & 0.226 (0.01) & 0.963 (0.00) \\
    \end{tabular}
    \caption{Loss functions comparison on three datasets using BERT-base architecture}
    \label{tab:losses-bert-base}
\end{table}

\section{Datasets}
\label{app:datasets}
\subsection{SNLI}

The dataset utilised in this study, initially introduced by \citet{bowman-etal-2015-large}  comprises a substantial collection of 570,000 pairs of human-authored English sentences. This dataset has a CC BY-SA 4.0 license. To ensure robustness, 10,000 pairs have been set aside for both testing and validation purposes. Within this corpus, the labels assigned to each sentence pair are evenly distributed among three distinct categories: entailment, neutral, and contradiction. In order to maintain consistent data sizes across all datasets and expedite the training process, a subset of 250,000 pairs was randomly sampled from the entire collection.

\subsection{Amazon Reviews}
The dataset utilised in this study, originally introduced by \citet{keung-etal-2020-multilingual}, was constructed by extracting customer reviews from a diverse range of product categories found on the Amazon marketplace. While reviews were collected in six different languages, only those composed in English were retained for training purposes. The dataset is structured for classification tasks and includes the corresponding star ratings, represented as integers spanning from 1 to 5. With a total of 210,000 samples, 5,000 samples were allocated separately for testing and validation purposes. Unlike the SNLI dataset, which underwent a sampling process, the entire dataset was utilised for training in this particular study.

\subsection{SST5}
The Stanford Sentiment Treebank (SST), originally introduced by \citet{socher-etal-2013-recursive}, serves as a dedicated corpus tailored specifically for sentiment analysis tasks. This dataset comes with a CC0 (public domain) license. What sets this corpus apart is its incorporation of parse trees, which enable comprehensive sentiment analysis at a granular level. The SST corpus consists of a meticulously curated collection of 12,000 sentences extracted from movie reviews, with each sentence undergoing thorough annotation by three human annotators. In the fine-grained variant of SST, known as SST-5, every individual phrase within the sentences is assigned a rating on a five-star scale. These ratings correspond to distinct sentiment categories, including negative, somewhat negative, neutral, somewhat positive, and positive, providing nuanced insights into the sentiment expressed within the sentences. 
\\ \\
Furthermore, for the few-shot learning scenario, we selected subsets consisting of 10\%, 25\%, and 50\% of the data in a randomised manner. This random selection process was repeated for multiple seed values, and the average results were reported to mitigate potential variations.

\section{Metrics}
\label{app:metrics}

\subsection{Nominal Metrics}

\subsubsection{F-1 Score}
The F1 score is a popular metric for evaluating classification models. It combines precision and recall to provide a single value that represents the model's overall performance. A good F1 score indicates the model's effectiveness in correctly classifying data points.

\[ F_1 = 2 \times \frac{{\text{{precision}} \times \text{{recall}}}}{{\text{{precision}} + \text{{recall}}}} \]

\subsection{Ordinal Metrics}

\subsubsection{MSE}

Mean Squared Error (MSE), quantifies the average squared difference between the predicted values and the corresponding actual values. A lower MSE signifies a superior alignment between the model's predictions and the ground truth values, thereby indicating heightened accuracy.
\[ \text{{MSE}} = \frac{1}{n} \sum_{i=1}^{n}(y_i - \hat{y_i})^2 \]

\subsubsection{MAE}
Mean Absolute Error (MAE), calculates the average absolute difference between the predicted values and the corresponding actual values. MAE offers a straightforward interpretation, representing the average magnitude of errors. A lower MAE value signifies a more accurate alignment between the model's predictions and the ground truth values, indicating a superior fit. 
\[ \text{{MAE}} = \frac{1}{n} \sum_{i=1}^{n}|y_i - \hat{y_i}| \]

\subsection{Off-by-k Accuracy}
Off-by-k accuracy is a metric used to evaluate the performance of a prediction model, particularly in the context of ranking or recommendation systems. It measures the percentage of predictions that are within $k$ positions of the correct prediction. 
\[ \text{OB}_k = 100 \times \frac{1}{S} \sum_{s=1}^{S} 1\{d(y_s, \hat{y}_s) \leq k\} \]

\begin{table*}[ht]
\centering
\small
\begin{tabular}{l|c } 
\hline
\textbf{Dataset} & \textbf{Label Verbaliser} \\ [0.5ex] 
\hline
SST-5 & \textit{indicates \{very negative / negative / neutral / positive / very positive\} sentiment } \\
AR & \textit{given \{very negative / negative / neutral / positive / very positive\} review }  \\
SNLI  & \textit{implies \{entailment / neutral / contradiction\} to}  \\
\hline
\end{tabular}
\caption{
\small
Verbalisers used for \textit{Entailment-style} approach. The texts inside \{\} show all the possible label descriptions available to construct the verbaliser.}
\label{table:prompt_details}
\end{table*}

\begin{table*}[ht]
\centering
\small
\begin{tabular}{l|c } 
\hline
\textbf{Dataset} & \textbf{Label Verbaliser} \\ [0.5ex] 
\hline
SST-5 & \textit{worse / bad / neutral / good / excellent} \\
AR & \textit{worse / bad / neutral / good / excellent} \\
SNLI  & \textit{yes / fair / no}  \\
\hline
\end{tabular}
\caption{
\small
Verbalisers used for \textit{Generative} approach.}
\label{table:prompt_details_gen}
\end{table*}

\section{Implementation Details}
\label{sec:train_detail}

We employed the \textit{BERT-base-uncased} model as the backbone for our vanilla CE, MLL, and entailment experiments. It consists of a total of 110M parameters including 12 encoder stacks, 12 attention heads, and a hidden state dimension of 768. This model was trained on lower-cased English text. To enable classification, we added a linear layer on top of the backbone model.

For the generative model experiments, we utilized \textit{GPT2-small} as the backbone model, which comprises 117M parameters, including 12 decoder stacks, 12 attention heads, and a hidden state dimension size of 768. We choose the \textit{small} variant to ensure fair comparison with its encoder-model counterpart.

For all the explicit loss-based experiments, we followed the standard supervised learning setup with the necessary modifications to loss function. For the Amazon Reviews (AR), 
and SST-5 datasets, the input to the model was the text/review and the corresponding class label was provided in a one-hot encoded format. For the SNLI dataset, we passed the input as \texttt{Premise [SEP] Hypothesis}.

For the entailment experiments, as mentioned in \S\ref{sec:implicit_entailment_intro}, we included one positive example and (K-1) negative examples. To address data imbalance, 
for AR, and SST-5, 
where there are five classes, we augmented an additional positive sample by randomly deleting 5\% of the text span in the input text. For SNLI, we created two negative examples corresponding to each positive sample. For all four datasets, we trained for 5 epochs using a learning rate of 5e-5, max sequence length 128, and batch size 32. Through experimentation, we determined the reported settings to yield the best results. Here, the input format to the model was - \texttt{verbaliser(label) [SEP] text} for 1-sentence tasks like AR, 
and SST-5. For 2-sentence task like SNLI, the input format was - \texttt{premise [SEP] verbaliser(label) [SEP] hypothesis}. 

For the generative model experiments, as mentioned in \S\ref{sec:implicit_decoder_intro}, we trained GPT2 on the language modeling task for 7 epochs using learning rate 1e-5, max sequence length 128, and batch size 32. Positioning the label verbaliser at the end of the text segment demonstrated the best results. The input is in the format - \texttt{text [SEP] verbaliser(label)} for 1-sentence tasks and \texttt{premise [SEP] hypothesis [SEP] verbaliser(label)} for 2-sentence task. For \texttt{verbaliser()} details refer to Table~\ref{table:prompt_details} and \ref{table:prompt_details_gen}.

For un-informative verbalisers, we used \textit{"cat"/"lion"/"zebra"/"dog"/"snake"} for SST-5, AR and \textit{"cat"/"lion"/"zebra"} for SNLI.

We also conducted additional experiments using the \textit{TinyBERT} model as the backbone to compare various loss functions. The TinyBERT model has a smaller number of 14.5M parameters and is more sensitive to hyperparameters. We trained this model using the same setup as the BERT-base-uncased experiments. The only difference was that we explored different variants of the OLL, MLL, and SOFT loss functions by tuning hyperparameters to compare them with other available loss functions such as CE, EMD, CORAL, etc.

For the Llama-Adapter based experiments, we keep the number of adapter layers at 30, the adaption verbaliser length at 10, max sequence length 512 to accommodate the instructions + verbaliser, batch size 4, weight decay 0.02, base learning rate 9e-3, warmup steps 2 and train it for 5 epochs. The instruction-verbaliser template is kept the same as Alpaca\footnote{\url{https://github.com/tloen/alpaca-lora}} format. 

All experiments were conducted with 5 different seeds on 8 Nvidia A100 GPUs in parallel. The reported results include the mean and standard deviations. The CE and MLL experiments took $\sim$2 hours 
for AR, and SNLI,
while SST-5 required $\sim$1 hour. In comparison, the entailment experiments took 3-5x longer to train since the dataset size was effectively increased by adding more negative \& positive samples. The generative model experiments, took $\sim$1 hour for 
AR, and SNLI
datasets with max sequence length 128, and  $\sim$30 mins for SST-5. All the reported training times are based on the full-data setting.

\section{PSR proofs}
\label{app:psr_proofs}
For a function to be proper scoring rule, it should attain its minimum value at the ground truth. For some losses, like CE and EMD, which are standard losses in ML/NLP, it is well established that they are satisfy by PSR. For others, we verify this property analytically by checking if loss goes to zero when we pass the ground truth information directly. Except for \textbf{SOFT} and \textbf{VS-SL} rest all other losses shown in Table~\ref{tab:summary_loss_properties} satisfy this PSR property. Below we give the overall idea on how to check for PSR condition.

\textbf{\textit{Cross-Entropy (CE):}} It is well established that CE belongs to PSR family \cite{gneiting2007strictly}. \\

\noindent \textbf{\textit{Ordinal Log Loss (OLL):}} OLL also belongs to the PSR family since the ground-truth one-hot encoding vector minimizes the loss. \\

\noindent \textbf{\textit{Multi-task log loss function (MLL):}} Since MLL is a weighted sum of CE and OLL, it can be said that the ground-truth one-hot encoding vector minimizes the MLL loss too since it is already established that both CE and OLL follow this property. \\

\noindent \textbf{\textit{SOFT labels (SOFT):}} The SOFT loss is similar to CE, computed with the soft ground labels due to which, the one hot encoded ground truth fails to minimise the loss. Hence SOFT loss doesn't belong to the PSR family.

\noindent \textbf{\textit{Earth Mover Distance (EMD) :}} The EMD loss for one particular arbitrary i will be - 

$$\text{EMD} = (\text{CDF}(\mathbb{I}(y_i)) - \text{CDF}(\hat{\bp_i}))^2$$
When the predicted probabilities $\hat{\bp_i}$ coincide with the one hot encoded ground truth $\mathbb{I}(y_i)$, loss value will be zero. Hence EMD does belong to the PSR family.

\noindent \textbf{\textit{Weighted Kappa Loss (WKL):}} The loss is defined as - 
\begin{align}
  \kappa = 1 - \frac{{\sum_{i, j} w_{ij} O_{ij}}}{{\sum_{i, j} w_{ij} E_{ij}}}
\end{align}

where $O_{ij}$ is the observed agreement between the annotators for class $i$ and class $j$, $E_{ij}$ is the expected agreement by chance, and $w_{ij}$ represents the weight assigned to each class pair.

To define the Weighted Kappa Loss function, we can formulate it as the negative value of the weighted kappa coefficient:
\begin{align}
 L_{\text{weighted\_kappa}} = -\kappa. 
\end{align}

If our classifier correctly predicts the ground truth our loss becomes zero, which guarantees that WKL is a PSR.

\section{Convexity proofs}
\label{app:convexity_proofs}

Most of the proofs are one-liners when we employ the standard results of convex optimization. We refer to \citet{boyd2004convex} for these standard results.

\textbf{\textit{Ordinal Log Loss (OLL):}} OLL is given by
\begin{align}
    -\sum_{i=1}^{N} \sum_{K: i_k \neq y_i} |i_k -y_i|^{\alpha} \log(1- \hat{p}_{i_k}). 
\end{align}

We will use the double derivative approach to prove the convexity. 
If we can prove the OLL to be convex for any one arbitrary i, then the whole function would also be convex(summation of convex functions is convex), so for an arbitrary i OLL is given by : 
    
\begin{align}
    -\sum_{K: i_k \neq y_i} |i_k -y_i|^{\alpha} \log(1- \hat{p}_{i_k}).
\end{align}

The  $|i_k -y_i|^{\alpha} $
     term is a positive constant with respect to $p_{i_k}$, so it won't affect the sign of double derivative and hence can be ignored for easier computations.
We now need to check the convexity of just : 
\begin{align}
    -\sum_{K: i_k \neq y_i} \log(1- \hat{p}_{i_k}). 
\end{align}

To prove the convexity of the given function, we will use the second derivative approach and try to prove that Hessian wrt $\{p_{i_k}\}$ terms is positive semi-definite (PSD) everywhere on the domain $[0,1]^{K-1}$ . For OLL, the Hessian will contain only diagonal terms and if we prove that each of these diagonal terms are positive, then we can say that Hessian is PSD and hence, OLL is convex. The diagonal term corresponding to $p_{i_k}$ is $\frac{1}{(1 - p_{i_k})^2}$ which is always non-negative.

\textbf{\textit{MLL:}} MLL can be viewed as a sum of two functions on the $K$-dimensional simplex $[0,1]^{K-1}$ i.e.

\begin{align}
    OLL - \log(1-\sum_{K:i_k \not = y_i} \hat{p_{i_k}})
\end{align}

Previously, we have seen that OLL is convex by proving the Hessian is PSD. Now, we also prove that the Hessian of $-\log(1-\sum_{K:i_k \neq y_i} \hat{p_{i_k}})$ is PSD on $[0,1]^{K-1}$ and hence, by the property that sum of two convex functions on the same domain is convex, we can say MLL is also convex. 

It is easy to verify Hessian of $-\log(1-\sum_{K:i_k \neq y_i} \hat{p_{i_k}})$ is $\frac{1}{(1-\sum_{K:i_k \neq y_i} \hat{p_{i_k}})^2}\mathbb{I}_{K-1 \times K-1}$ where $\mathbb{I}_{K-1 \times K-1
}$ is an identity matrix of dimension $K-1 \times K-1$. For the Hessian to be PSD, all diagonal values must be non-negative and it's true in this case. Hence the function $-\log(1-\sum_{K:i_k \neq y_i} \hat{p_{i_k}})$ is convex along with the OLL being convex in the same domain, which makes MLL a convex loss function.

\textbf{\textit{SOFT labels (SOFT):}}

The SOFT loss is given by :
\begin{align}
    -\sum_{i=1}^{N} \sum_{k'} p_{i_{k' }}^{\text {soft }}\log(\hat{p}_{i_{k'}}) 
\end{align}

where - 

\begin{align}
    p_{i_k}^{\text {soft }}=\frac{\exp (-\beta |i_k - y_i|)}{\sum_{k'} \exp (-\beta |i_{k'}- y_i|)} 
\end{align}

The loss function bears resemblance to the Cross-Entropy (CE) loss, with the distinction that the ground truth one-hot encoded label is replaced by softlabels. The term $p_{i_k}^{\text {soft}}$ represents a positive constant concerning the differentiating variable and does not affect the sign of the double derivative. The convexity of the remaining function can be demonstrated in a similar manner as the CE loss. Alternatively, the convex nature of CE implies the convexity of the SOFT loss, and the convexity of CE has been previously established.

\textbf{\textit{Earth Mover Distance (EMD)}} 
EMD is also known as Wassertein distance and is well known to be a convex loss on the probabilities \cite{merigot2021non}.

\newpage

\end{document}